\documentclass{article}

\usepackage{microtype}
\usepackage{graphicx}
\usepackage{subfigure}
\usepackage{booktabs} 

\usepackage{hyperref}

\usepackage[accepted]{icml2024}

\usepackage{siunitx}  
\usepackage{multicol}
\usepackage{amsmath}
\usepackage{amsfonts}
\usepackage{amssymb}
\usepackage{amsthm}
\usepackage{eurosym}
\usepackage{bm}
\usepackage{natbib}
\usepackage{multirow}
\usepackage{subcaption}
\usepackage{soul}

\usepackage[capitalize,noabbrev]{cleveref}

\usepackage{enumitem}
\setlist[itemize,enumerate]{topsep=4pt,itemsep=0pt,leftmargin=*}

\usepackage{comment}

\theoremstyle{plain}

\usepackage{mdframed}
\definecolor{theoremcolor}{rgb}
{0.97, 0.97, 0.9}
\definecolor{examplecolor}{rgb}{1, 1, 1.0}
\mdfsetup{
    backgroundcolor=theoremcolor,
    linewidth=0pt,
}

\newmdtheoremenv[linewidth=0pt,innerleftmargin=4pt,innerrightmargin=4pt]{definition}{Definition}
\newmdtheoremenv[linewidth=0pt,innerleftmargin=4pt,innerrightmargin=4pt]{proposition}{Proposition}
\newmdtheoremenv[linewidth=0pt,innerleftmargin=0pt,innerrightmargin=0pt,backgroundcolor=examplecolor]{example}{Example}
\newmdtheoremenv{corollary}{Corollary}
\newmdtheoremenv{theorem}{Theorem}
\newmdtheoremenv{lemma}{Lemma}

\DeclareMathOperator*{\argmax}{argmax}

\usepackage[textsize=tiny]{todonotes}

\icmltitlerunning{Sparse and Structured Hopfield Networks}

\begin{document}

\twocolumn[
\icmltitle{Sparse and Structured Hopfield Networks}



\icmlsetsymbol{equal}{*}

\begin{icmlauthorlist}
\icmlauthor{Saul Santos}{ist,it}
\icmlauthor{Vlad Niculae}{uva}
\icmlauthor{Daniel McNamee}{cnp}
\icmlauthor{André F. T. Martins}{ist,it,unbabel}
\end{icmlauthorlist}

\icmlaffiliation{ist}{Instituto Superior Técnico, Universidade de Lisboa, Lisbon, Portugal}
\icmlaffiliation{it}{Instituto de Telecomunicações, Lisbon, Portugal}
\icmlaffiliation{uva}{Language Technology Lab, University of Amsterdam, The Netherlands}
\icmlaffiliation{cnp}{Champalimaud Research, Lisbon, Portugal}
\icmlaffiliation{unbabel}{Unbabel, Lisbon, Portugal}

\icmlcorrespondingauthor{Saul Santos}{saul.r.santos@tecnico.ulisboa.pt}

\icmlkeywords{Machine Learning, ICML}

\vskip 0.3in
]



\printAffiliationsAndNotice{\icmlEqualContribution} 

\begin{abstract}
Modern Hopfield networks have enjoyed recent interest due to their 
connection to
attention in transformers. 
Our paper provides a unified framework for sparse Hopfield networks by establishing a link with Fenchel-Young losses. 
The result is a new family of Hopfield-Fenchel-Young energies whose update rules are end-to-end differentiable \textit{sparse} transformations. 
We reveal a connection between loss margins, sparsity, and \textit{exact} memory retrieval. 
We further extend this framework to \textit{structured} Hopfield networks via the SparseMAP transformation, which can retrieve pattern associations instead of a single pattern. 
Experiments on multiple instance learning and text rationalization demonstrate the usefulness of our approach. 

\end{abstract}

\section{Introduction}


Hopfield networks are a kind of biologically-plausible neural network
with
associative memory capabilities \citep{hopfield1982neural}. Their attractor dynamics makes them suitable for modeling the retrieval of episodic memories in humans and animals \citep{tyulmankov2021biological,whittington2021relating}.
The limited storage capacity of 
classical 
Hopfield networks
was
recently overcome through
new energy functions and continuous state patterns \citep{krotov2016dense,
demircigil2017model,ramsauer2020hopfield}, leading to exponential storage
capacities and sparking renewed interest in ``modern'' Hopfield networks.
In particular, \citet{ramsauer2020hopfield} revealed
striking connections to transformer attention,
but their model is incapable of exact retrieval and 
may require a low temperature to avoid metastable states (states which mix multiple input patterns).

In this paper, we bridge this gap by making a connection between Hopfield energies and \textbf{Fenchel-Young losses} \citep{blondel2020learning}. We extend the energy functions of 
\citet{ramsauer2020hopfield} and \citet{hu2023sparse} 
to a wider family
induced by generalized
entropies.
The minimization of these energy functions leads to \textbf{sparse update rules} which include as particular cases 
$\alpha$-entmax \citep{peters2019sparse}, $\alpha$-normmax \citep{blondel2020learning},   
and {SparseMAP} \cite{niculae2018sparsemap}, where a structural constraint 
ensures the retrieval of $k$ patterns. Unlike
\citet{ramsauer2020hopfield}'s Hopfield layers, our update rules pave the way for \textbf{exact retrieval}, leading to the emergence of sparse association among patterns while ensuring end-to-end differentiability. 

This endeavour aligns with the strong neurobiological motivation to seek new Hopfield energies capable of \textbf{sparse} and \textbf{structured} retrieval.
Indeed, sparse neural activity patterns forming structured representations underpin core principles of cortical computation \cite{SimoncelliOlhausen2001,Tse2007,Palm2013}.
With respect to memory formation circuits, the sparse firing of neurons in the dentate gyrus (DG), a distinguished region within the hippocampal network, underpins its proposed role in pattern separation during memory storage \citep{Yassa2011,Severa2017}. Evidence suggests that such sparsified 
activity aids in minimizing interference,  
however an integrative theoretical account linking sparse coding and attractor network functionality to clarify these empirical observations is lacking
\citep{Leutgeb2007a,Neunuebel2014}.
\begin{figure*}[t]
    \centering
    \includegraphics[width=1\textwidth]{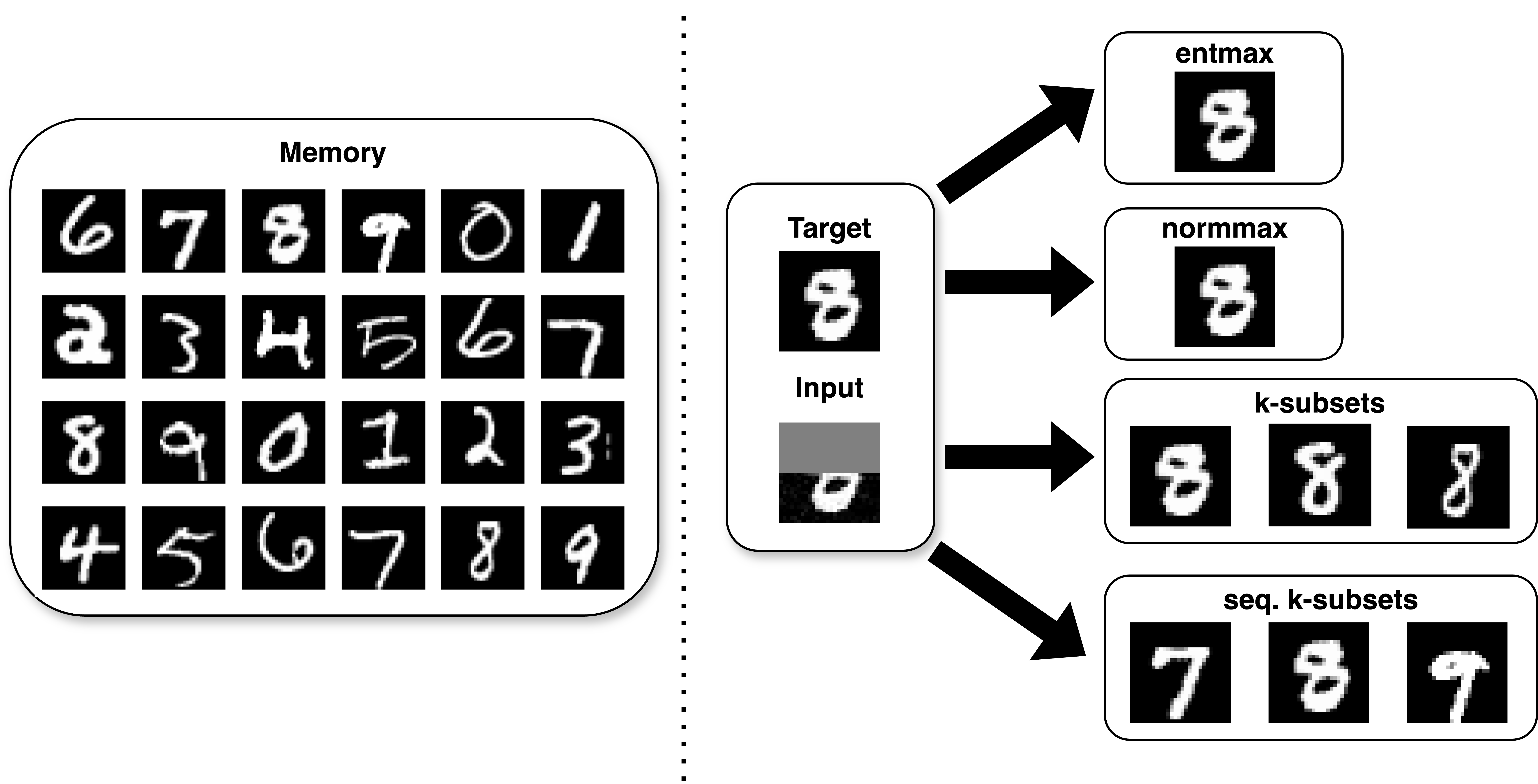}
    \caption{Overview of the Hopfield networks proposed in this paper: sparse transformations (entmax and normmax) aim to retrieve the closest pattern to the query, and they have exact retrieval guarantees. Structured variants find pattern associations. The $k$-subsets transformation returns a mixture of the top-$k$ patterns, and sequential $k$-subsets favors contiguous retrieval.}
    \label{fig:overview}
\end{figure*}
Our main contributions are:
\begin{itemize}
    \item We introduce \textbf{Hopfield-Fenchel-Young} energy functions as a  generalization of modern Hopfield networks (\S\ref{sec:hfy_definition}). 
    \item We leverage  properties of Fenchel-Young losses which relate \textbf{sparsity} and \textbf{margins} to obtain new theoretical results for 
    \text{exact} memory retrieval, proving exponential storage capacity in a stricter sense than prior work (\S\ref{sec:hfy_margins}). 
    \item We propose  new \textbf{structured} Hopfield networks via the SparseMAP transformation,  which return \textbf{pattern associations} instead of single patterns. We show that SparseMAP has a structured margin and 
    provide sufficient conditions for exact retrieval of pattern associations (\S\ref{sec:structured}). 
\end{itemize}
Experiments on synthetic and real-world tasks (multiple instance learning and text rationalization) showcase the usefulness of our proposed models using various kinds of sparse and structured transformations (\S\ref{sec:experiments}). An overview of the utility of our methods can be seen in Fig. \ref{fig:overview}.%
\footnote{Our code is available on \url{https://github.com/deep-spin/SSHN}} %
\paragraph{Notation.} We denote by $\triangle_N$ the $(N-1)\textsuperscript{th}$-dimensional probability simplex, $\triangle_N := \{\bm{p} \in \mathbb{R}^N : \bm{p} \ge \mathbf{0}, \, \mathbf{1}^\top \bm{p} = 1\}$. 
The convex hull of a set $\mathcal{Y} \subseteq \mathbb{R}^M$ is $\mathrm{conv}(\mathcal{Y}) := \{\sum_{i=1}^N p_i \bm{y}_i \,:\, \bm{p} \in \triangle_N, \, \bm{y}_1, ..., \bm{y}_N \in \mathcal{Y}, \, N \in \mathbb{N}\}$. 
We have $\triangle_N = \mathrm{conv}(\{\bm{e}_1, ..., \bm{e}_N\})$, where $\bm{e}_i \in \mathbb{R}^N$ is the $i$\textsuperscript{th} basis (one-hot) vector. 
Given a convex function $\Omega: \mathbb{R}^N \rightarrow \bar{\mathbb{R}}$, where $\bar{\mathbb{R}} = {\mathbb{R}} \cup \{+\infty\}$, we denote its domain by $\mathrm{dom}(\Omega) := \{\bm{y} \in \mathbb{R}^N \,:\, \Omega(\bm{y}) < +\infty\}$ and its Fenchel conjugate by $\Omega^*(\bm{\theta}) = \sup_{\bm{y} \in \mathbb{R}^N} \bm{y}^\top \bm{\theta} - \Omega(\bm{y})$. 
We denote by $I_\mathcal{C}$ the indicator function of a convex set $\mathcal{C}$, defined as $I_\mathcal{C}(\bm{y}) = 0$ if $\bm{y} \in \mathcal{C}$, and $I_\mathcal{C}(\bm{y}) = +\infty$ otherwise. 
\section{Background}\label{sec:background}

\subsection{Hopfield Networks}

Let $\bm{X} \in \mathbb{R}^{N \times D}$ be a matrix whose rows hold a set of examples $\bm{x}_1, ..., \bm{x}_N \in \mathbb{R}^D$ (``memory patterns'') and let $\bm{q}^{(0)} \in \mathbb{R}^D$ be a query vector
(or ``state pattern''). 
Hopfield networks iteratively
update $\bm{q}^{(t)} \mapsto \bm{q}^{(t+1)}$ for $t\in \{0, 1, ...\}$ according to a certain rule, eventually converging to a fixed point attractor state $\bm{q}^*$ which either
corresponds
to one of the memorized examples, or to a mixture
thereof.
This update rule correspond to the minimization of an energy function, 
which for classic Hopfield networks \citep{hopfield1982neural} takes the form 
$E(\bm{q}) = -\frac{1}{2} \bm{q}^\top \bm{W} \bm{q}$, with $\bm{q} \in \{\pm 1\}^D$ and $\bm{W} = \bm{X}^\top\bm{X} \in \mathbb{R}^{D \times D}$, leading to the update rule $\bm{q}^{(t+1)} = \mathrm{sign}(\bm{W}\bm{q}^{(t)})$. A limitation of this classical network is that it has only $N=\mathcal{O}(D)$ memory storage capacity, above which patterns start to interfere \citep{amit1985storing,mceliece1987capacity}. 

Recent work 
sidesteps
this limitation through alternative energy functions \citep{krotov2016dense,demircigil2017model}, paving the way for a class of models known as ``modern Hopfield networks'' with superlinear (often exponential) memory capacity. In \citet{ramsauer2020hopfield}, $\bm{q} \in \mathbb{R}^D$ is continuous and the following energy is used:
\begin{align}
    \label{eq:energy_hopfield}
        E(\bm{q}) = -\frac{1}{\beta} \log\sum_{i=1}^N \exp(\beta \bm{x}_i^\top \bm{q}) + \frac{1}{2} \|\bm{q}\|^2 + \mathrm{const.}
\end{align}

\citet{ramsauer2020hopfield} revealed an interesting relation between the updates in this modern Hopfield network and the attention layers in transformers. 
Namely, the minimization of the energy \eqref{eq:energy_hopfield} using the concave-convex procedure (CCCP; \citealt{yuille2003concave}) leads to the update rule:
\begin{equation}\label{eq:update_rule}
    \begin{aligned}
        \bm{q}^{(t+1)} &= \bm{X}^\top \mathrm{softmax}(\beta \bm{X} \bm{q}^{(t)}).
    \end{aligned}
\end{equation}
When $\beta = \frac{1}{\sqrt{D}}$, each update matches
the computation performed in the attention layer of a transformer with a single attention head and with identity projection matrices.
This triggered interest in developing variants of Hopfield layers which can be used as drop-in replacements for multi-head attention layers \citep{hoover2023energy}.

While \citet{ramsauer2020hopfield} have derived useful theoretical properties of these networks (including their exponential storage capacity under a relaxed notion of retrieval), the use of
softmax
in 
\eqref{eq:update_rule} prevents exact convergence and may lead to undesirable metastable states. 
Recently (and closely related to our work), \citet{hu2023sparse} 
introduced sparse Hopfield networks and 
proved favorable retrieval properties,  but still under an approximate sense, where attractors are close but distinct from the stored patterns. 
We overcome these drawbacks in our paper, where we provide a unified treatment of 
sparse and structured  Hopfield networks along with theoretical analysis showing that exact retrieval is possible without sacrificing exponential storage capacity. 
\subsection{Sparse Transformations and Fenchel-Young Losses}

Our construction and results flow from the properties of Fenchel-Young losses, which we next review.

Given a strictly convex function $\Omega: \mathbb{R}^N \rightarrow \bar{\mathbb{R}}$ with domain $\mathrm{dom}(\Omega)$,
the 
$\Omega$-regularized argmax transformation \citep{niculae2017regularized}, 
$\hat{\bm{y}}_\Omega : \mathbb{R}^N \rightarrow \mathrm{dom}(\Omega)$, is:
\begin{align}\label{eq:rpm}
    \hat{\bm{y}}_\Omega(\bm{\theta}) := \nabla \Omega^*(\bm{\theta}) = \argmax_{\bm{y} \in \mathrm{dom}(\Omega)} \bm{\theta}^\top \bm{y} - \Omega(\bm{y}).
\end{align}
Let us assume for now that $\mathrm{dom}(\Omega) = \triangle_N$ (the probability simplex). 
One 
instance of
\eqref{eq:rpm} is the \textbf{softmax} transformation, obtained when the regularizer is the Shannon negentropy, $\Omega(\bm{y}) = \sum_{i=1}^N y_i \log y_i + I_{\triangle_N}(\bm{y})$. 
Another 
instance
is the \textbf{sparsemax} transformation, obtained when $\Omega(\bm{y}) = \frac{1}{2}\|\bm{y}\|^2 + I_{\triangle_N}(\bm{y})$ \citep{martins2016softmax}.
The sparsemax transformation is 
the Euclidean projection onto the probability simplex.
Softmax and sparsemax are both particular cases of \textbf{$\alpha$-entmax transformations} \citep{peters2019sparse}, parametrized by a scalar $\alpha \ge 0$ (called the entropic index), which corresponds to the following choice of regularizer, called the \textbf{Tsallis $\alpha$-negentropy} \citep{tsallis1988possible}:
\begin{align}\label{eq:tsallis}
    \Omega^T_\alpha(\bm{y}) = \frac{-1 + \|\bm{y}\|_\alpha^\alpha}{\alpha(\alpha-1)} + I_{\triangle_N}(\bm{y}).
\end{align}
When $\alpha \rightarrow 1$,
the Tsallis $\alpha$-negentropy
\(\Omega_\alpha^T\)
becomes Shannon's negentropy, leading to the softmax transformation, and when $\alpha=2$, it becomes the $\ell_2$-norm (up to a constant) and we recover the sparsemax.
\begin{figure}
    \centering
    \includegraphics[width=\columnwidth]{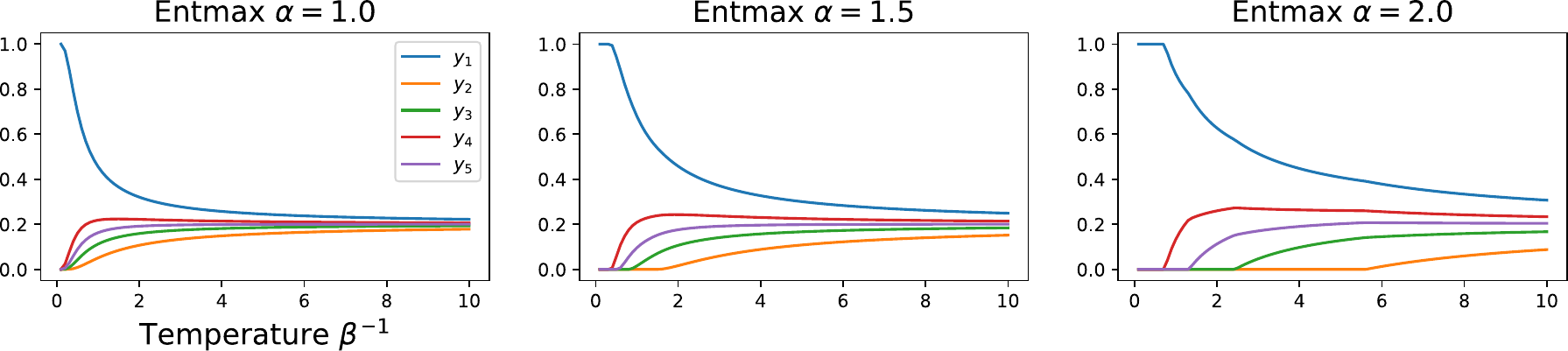}\\
    \includegraphics[width=\columnwidth]{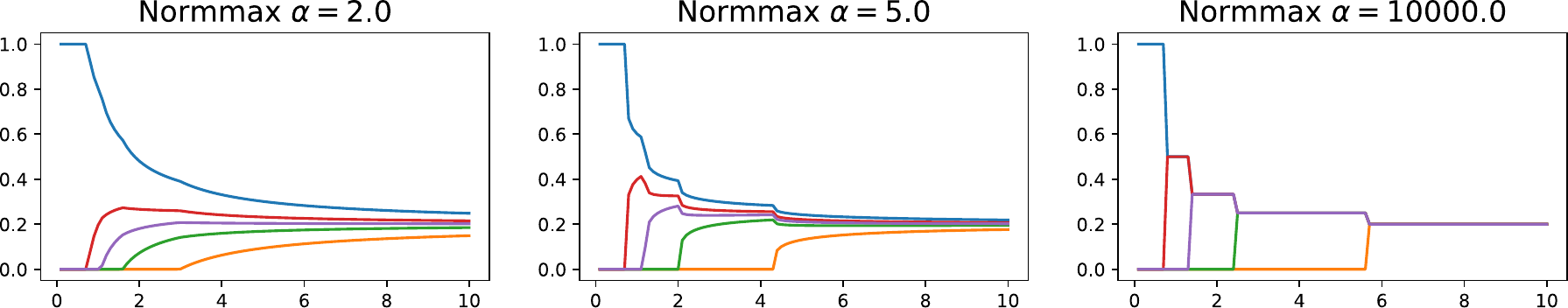}\\
    \vspace{0.3cm}
    \includegraphics[width=\columnwidth]{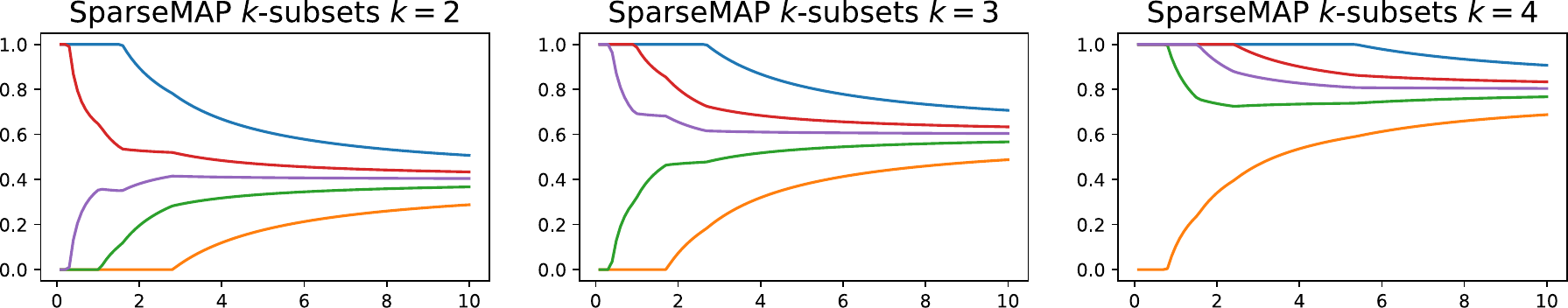}
    \caption{Sparse and structured transformations used in this paper and their regularization path. In each plot, we show $\hat{\bm{y}}_\Omega(\beta \bm{\theta}) = \hat{\bm{y}}_{\beta^{-1}\Omega}(\bm{\theta})$ as a function of the temperature $\beta^{-1}$ where $\bm{\theta} = [1.0716, -1.1221, -0.3288, 0.3368, 0.0425]^\top$. Additional examples can be found in App. \ref{sec:SST_App}.}
    \label{fig:sparse_transformations}
\end{figure} 
Another example is the  \textbf{norm $\alpha$-negentropy} \citep[\S4.3]{blondel2020learning}, 
\begin{align}\label{eq:norm_entropy}
    \Omega^{N}_{\alpha}(\bm{y}) = -1 + \|\bm{y}\|_\alpha + I_{\triangle_N}(\bm{y}), 
\end{align}
which, when $\alpha \rightarrow +\infty$, is called the Berger-Parker dominance index \citep{may1975patterns}. We call the resulting transformation \textbf{$\alpha$-normmax}.  
While the Tsallis and norm negentropies have similar expressions and the resulting transformations both tend to be sparse, they have important differences, as suggested in Fig.~\ref{fig:sparse_transformations}: 
normmax favors distributions closer to uniform over the selected support.

The \textbf{$\Omega$-Fenchel-Young loss} 
\citep{blondel2020learning} is 
\begin{align}\label{eq:fy_loss}
    L_{\Omega}(\bm{\theta}, \bm{y}) := \Omega(\bm{y}) + \Omega^*(\bm{\theta}) - \bm{\theta}^\top \bm{y}.
\end{align}
When $\Omega$ is Shannon's negentropy, 
we have $\Omega^*(\bm{\theta}) = \log\sum_{i=1}^N\exp(\theta_i)$,
and $L_\Omega$ is the \textbf{cross-entropy loss}, up to a constant.  
Intuitively, Fenchel-Young losses quantify how ``compatible'' a score vector $\bm{\theta} \in \mathbb{R}^N$ (e.g., logits) is to a desired target $\bm{y} \in \mathrm{dom}(\Omega)$ (e.g., a probability vector). 

Fenchel-Young losses have 
relevant
properties \citep[Prop.~2]{blondel2020learning}: 
(i) 
they are non-negative, $L_\Omega(\bm{\theta}, \bm{y}) \ge 0$, with equality iff $\bm{y} = \hat{\bm{y}}_\Omega(\bm{\theta})$; 
(ii) they are convex on $\bm{\theta}$ and their gradient is 
$\nabla_{\bm{\theta}} L_\Omega(\bm{\theta}, \bm{y}) = -\bm{y} + \hat{\bm{y}}_\Omega(\bm{\theta})$.
%
%
But, most importantly, they have well-studied \textbf{margin} properties,
which, as we shall see, will play a pivotal role in this work. 

\begin{definition}[Margin]\label{def:margin} A loss function $L(\bm{\theta}; \bm{y})$ has a \textbf{margin} if there exists a
finite $m \ge 0$ such that
\begin{align}\label{eq:margin}
\begin{split}
\forall i \in [N], \quad
    L(\bm{\theta}, \bm{e}_i) = 0     &\Longleftrightarrow {\theta}_i - \max_{j \ne i} {\theta}_j \ge m.
\end{split}
\end{align}
The smallest such $m$ is called the margin of $L$. 
If $L_\Omega$ is a Fenchel-Young loss, 
\eqref{eq:margin} is equivalent to $\hat{\bm{y}}_{\Omega}(\bm{\theta}) = \bm{e}_i$. 
\end{definition} 
A famous example of a loss with a margin of 1 is the hinge loss of support vector machines. 
On the other hand, the cross-entropy loss does not have a margin. 
\citet[Prop. 7]{blondel2020learning} have shown that Tsallis negentropies $\Omega^T_\alpha$ with $\alpha > 1$ have a margin of $m = (\alpha-1)^{-1}$, and that norm-entropies $\Omega^N_\alpha$ with $\alpha > 1$ have a margin $m=1$, independently of $\alpha$. We will use these facts in the sequel. 

\section{Sparse Hopfield-Fenchel-Young Energies}\label{sec:probabilistic}

We now use Fenchel-Young losses \eqref{eq:fy_loss} to define a new class of energy functions for modern Hopfield networks. 

\subsection{Definition and update rule}\label{sec:hfy_definition}

We start by assuming that the regularizer $\Omega$ has domain $\mathrm{dom}(\Omega) = \triangle_N$ and that it is a \textbf{generalized negentropy}, i.e., null when \(\bm{y}\) is a one-hot vector, strictly convex, and permutation-invariant (see Appendix~\ref{sec:generalized_negent} for details). 
These conditions imply that $\Omega \le 0$ and that $\Omega(\bm{y})$ is minimized when $\bm{y}=\mathbf{1}/N$ is the uniform distribution \citep[Prop.~4]{blondel2020learning}.
Tsallis negentropies \eqref{eq:tsallis} for $\alpha \ge 1$ and  norm negentropies \eqref{eq:norm_entropy} for $\alpha>1$ both satisfy these properties. 

We define the \textbf{Hopfield-Fenchel-Young (HFY) energy} as
\begin{align}\label{eq:hfy_energy}
    E(\bm{q}) 
    &= \underbrace{-\beta^{-1} L_\Omega(\beta \bm{X} \bm{q}; \mathbf{1}/{N})}_{E_{\mathrm{concave}}(\bm{q})} \nonumber\\ 
    &+ \underbrace{\frac{1}{2} \|\bm{q} - \bm{\mu}_{\bm{X}}\|^2 + \frac{1}{2}(M^2 - \|\bm{\mu}_{\bm{X}}\|^2)}_{{E_{\mathrm{convex}}(\bm{q})}},
\end{align}
where $\bm{\mu}_{\bm{X}} := \bm{X}^\top \mathbf{1}/N \in \mathbb{R}^D$ is the empirical mean of the patterns, and $M := \max_i \|\bm{x}_i\|$. 
This energy extends that of \eqref{eq:energy_hopfield}, which is recovered when $\Omega$ is Shannon's negentropy. 
The concavity of $E_{\mathrm{concave}}$ holds from the convexity of Fenchel-Young losses on its first argument and from the fact that composition of a convex function with an affine map is convex. 
$E_{\mathrm{convex}}$ is convex because it is quadratic.%
\footnote{Up to constants, for this choice of $\Omega$ our convex-concave decomposition is the same as 
\citet{ramsauer2020hopfield}.
} %

These two
terms compete when minimizing the energy 
\eqref{eq:hfy_energy}: 
\begin{itemize}
    \item Minimizing $E_{\mathrm{concave}}$ is equivalent to \textit{maximizing} $L_{\Omega}(\beta \bm{X} \bm{q}; \mathbf{1}/{N})$, 
    which pushes
    $\bm{q}$ to be as far as possible from a uniform average and closer to a single pattern.
    \item Minimizing $E_{\mathrm{convex}}$ serves as a proximity regularization, encouraging the state pattern $\bm{q}$ to stay close to $\bm{\mu}_{\bm{X}}$.
\end{itemize} 
The next result, proved in App.~\ref{sec:proof_prop_bounds_cccp}, establishes bounds and derives the Hopfield update rule for energy \eqref{eq:hfy_energy}, generalizing \citet[Lemma A.1, Theorem A.1]{ramsauer2020hopfield}.

\begin{proposition}[Update rule of HFY energies]\label{prop:bounds_cccp}
    Let the query $\bm{q}$ be in the convex hull of the rows of $\bm{X}$, i.e., $\bm{q} = \bm{X}^\top \bm{y}$ for some $\bm{y} \in \triangle_N$. Then, the energy \eqref{eq:hfy_energy} satisfies
\(0 \le E(\bm{q}) \le \min\left\{2M^2, \,\, -\beta^{-1}\Omega(\mathbf{1}/N) + \frac{1}{2}M^2\right\}\).
Furthermore, minimizing \eqref{eq:hfy_energy} with the CCCP algorithm \citep{yuille2003concave} leads to the updates:
\begin{align}\label{eq:updates_hfy}
    \bm{q}^{(t+1)} = \bm{X}^\top \hat{\bm{y}}_\Omega(\beta \bm{X} \bm{q}^{(t)}).
\end{align}
\end{proposition}

In particular, when $\Omega=\Omega^T_\alpha$ (the Tsallis $\alpha$-negentropy \eqref{eq:tsallis}), the update \eqref{eq:updates_hfy} corresponds to the adaptively sparse transformer of \citet{correia2019adaptively}. 
The $\alpha$-entmax transformation can be computed in linear time for $\alpha \in \{1, 1.5, 2\}$ and for other values of $\alpha$ an efficient bisection algorithm was proposed by \citet{peters2019sparse}. 
The case $\alpha=2$ (sparsemax) corresponds to the sparse modern Hopfield network recently proposed by \citet{hu2023sparse}. 

When $\Omega=\Omega^N_\alpha$ (the norm $\alpha$-negentropy \eqref{eq:norm_entropy}), we obtain the $\alpha$-normmax transformation. This transformation is harder to compute since $\Omega^N_\alpha$ is not separable, but we derive in App.~\ref{sec:normmax} an efficient bisection algorithm which works for any $\alpha>1$. 




\subsection{Margins, sparsity, and exact retrieval}\label{sec:hfy_margins}

Prior work on modern Hopfield networks \citep[Def.~1]{ramsauer2020hopfield} 
defines pattern storage and retrieval in an {\it approximate} sense: they assume a small neighbourhood around each pattern $\bm{x}_i$ containing an attractor $\bm{x}_i^*$, such that if the initial query $\bm{q}^{(0)}$ is close enough, the Hopfield updates will converge to $\bm{x}_i^*$, leading to a retrieval error of $\|\bm{x}_i^* - \bm{x}_i\|$. For this error to be small, a large $\beta$ may be necessary. 

We consider here a stronger definition of \textbf{exact retrieval}, where the attractors \textit{coincide} with the actual patterns (rather than being nearby). 
Our main result is that 
\textbf{zero retrieval error} is possible in HFY networks as long as the corresponding Fenchel-Young loss has a \textbf{margin} 
(Def.~\ref{def:margin}). 
Given that $\hat{\bm{y}}_\Omega$ being a sparse transformation is a sufficient condition for $L_\Omega$ having a margin \citep[Prop.~6]{blondel2020learning}, this is a general statement about sparse transformations.%
\footnote{At first sight, this might seem a surprising result,  given that both queries and patterns are continuous. The reason why exact convergence is possible hinges crucially on sparsity.} 

\begin{definition}[Exact retrieval]\label{def:exact_retrieval}
    A pattern $\bm{x}_i$ is \textbf{exactly retrieved} for  query $\bm{q}^{(0)}$ iff there is a finite number of steps $T$ such that  iterating \eqref{eq:updates_hfy} 
    leads to $\bm{q}^{(T')} = \bm{x}_i$ $\forall T' \ge T$. 
\end{definition}

The following result gives sufficient conditions for exact retrieval with $T=1$ given that patterns are well separated and that the query is sufficiently close to the retrieved pattern. 
It establishes the exact \textbf{autoassociative} property of HFY networks: if all patterns are slightly perturbed, the Hopfield dynamics are able to recover the original patterns exactly.  
Following \citet[Def.~2]{ramsauer2020hopfield}, we define the separation of pattern $\bm{x}_i$ from data as $\Delta_i = \bm{x}_i^\top \bm{x}_i - \max_{j \ne i} \bm{x}_i^\top \bm{x}_j$.

\begin{proposition}[Exact retrieval in a single iteration]\label{prop:separation}
    Assume $L_\Omega$ has margin $m$, 
    and let $\bm{x}_i$ be a  pattern outside the convex hull of the other patterns.
    Then, $\bm{x}_i$ is a stationary point of the energy \eqref{eq:hfy_energy} iff $\Delta_i \ge m{\beta^{-1}}$. 
    In addition, if the initial query $\bm{q}^{(0)}$ satisfies ${\bm{q}^{(0)}}^\top (\bm{x}_i -\bm{x}_j) \ge m{\beta^{-1}}$ for all $j \ne i$, then the update rule \eqref{eq:updates_hfy} converges to $\bm{x}_i$ exactly in one iteration. Moreover, if the patterns are normalized, $\|\bm{x}_i\| = M$ for all $i$, and well-separated with $\Delta_i \ge m{\beta^{-1}} + 2M\epsilon$, then any $\bm{q}^{(0)}$ $\epsilon$-close to $\bm{x}_i$ ($\|\bm{q}^{(0)} - \bm{x}_i\| \le \epsilon$) will converge to $\bm{x}_i$ in one iteration. 
\end{proposition}



The proof is in Appendix~\ref{sec:proof_prop_separation}. 
For the Tsallis negentropy case $\Omega = \Omega^T_\alpha$ with $\alpha>1$ (the sparse case), 
we have $m = (\alpha-1)^{-1}$ (cf.~Def.~\ref{def:margin}), leading to the condition $\Delta_i \ge \frac{1}{(\alpha-1)\beta}$. 
This result is stronger than that of \citet{ramsauer2020hopfield} for their energy (which is ours for $\alpha=1$), according to which memory patterns are only $\epsilon$-close to stationary points, where a small $\epsilon = 
\mathcal{O}(\exp(-\beta))$ requires a low temperature (large $\beta$). 
It is also stronger than the retrieval error bound recently derived by \citet[Theorem 2.2]{hu2023sparse} for the case $\alpha=2$, which has an additive term involving $M$ and therefore does not provide conditions for exact retrieval. 

For the normmax negentropy case $\Omega = \Omega^N_\alpha$ with $\alpha>1$, we have $m=1$, so the condition above becomes $\Delta_i \ge \frac{1}{\beta}$.


Given that exact retrieval is a stricter definition, one may wonder whether requiring it sacrifices storage capacity. 
Reassuringly, the next result, inspired but stronger than \citet[Theorem~A.3]{ramsauer2020hopfield}, shows that HFY networks with exact retrieval also have exponential storage capacity.  

\begin{proposition}[Storage capacity with exact retrieval]\label{prop:storage}
Assume patterns are placed equidistantly on the sphere of radius $M$. 
The HFY network can store and exactly retrieve $N = \mathcal{O}((2/\sqrt{3})^D)$ 
patterns in one iteration under a $\epsilon$-perturbation as long as 
$M^2 > 2m{\beta^{-1}}$ and
\begin{align}
    \epsilon \le \frac{M}{4} - \frac{m}{2\beta M}. 
\end{align}
Assume patterns are randomly placed on the sphere with uniform distribution. 
Then, with probability $1-p$, the HFY network can store and exactly retrieve $N = \mathcal{O}(\sqrt{p} \zeta^{\frac{D-1}{2}})$ 
patterns in one iteration under a $\epsilon$-perturbation 
if
\begin{equation}
    \epsilon \le \frac{M}{2} \left(1 - \cos \frac{1}{\zeta}\right) - \frac{m}{2\beta M}.
\end{equation}
\end{proposition}
The proof is in Appendix~\ref{sec:proof_prop_storage}.

\section{Structured Hopfield Networks}\label{sec:structured}


In \S\ref{sec:probabilistic}, we considered the case where $\bm{y} \in \mathrm{dom}(\Omega) = \triangle_N$, the scenario studied by \citet{ramsauer2020hopfield} and \citet{hu2023sparse}. 
We now take one step further and consider the more general \textbf{structured} case, where $\bm{y}$ is a vector of ``marginals'' associated to some given structured set. 
This structure can reflect \textbf{pattern associations} that we might want to induce when querying the Hopfield network with $\bm{q}^{(0)}$. 
Possible structures include \textbf{$k$-subsets} of memory patterns, potentially leveraging \textbf{sequential memory structure}, tree structures,  matchings, etc. 
In these cases, the set of pattern associations we can form is combinatorial, hence it can be considerably larger 
than the number $N$ of memory patterns.  

\subsection{Unary scores and structured constraints}\label{sec:structured_unary}

We consider first a simpler scenario where there is a predefined set of structures $\mathcal{Y} \subseteq \{0, 1\}^N$ and $N$ unary scores, one for each memory pattern. We show in \S\ref{sec:structured_high_order} how this framework can be generalized for higher-order interactions modeling soft interactions among patterns. 

Let $\mathcal{Y} \subseteq \{0, 1\}^N$ be a set of binary vectors indicating the underlying set of structures, and let $\mathrm{conv}(\mathcal{Y}) \subseteq [0,1]^N$ denote its convex hull, called the \textbf{marginal polytope} associated to the structured set $\mathcal{Y}$ \citep{wainwright2008graphical}. 

\begin{example}[$k$-subsets]\label{ex:ksubsets}
We may be interested in retrieving subsets of $k$ patterns, e.g., to take into account a $k$-ary relation among patterns or to perform top-$k$ retrieval. In this case, we define $\mathcal{Y} = \{\bm{y} \in \{0, 1\}^N \,:\, \mathbf{1}^\top \bm{y} = k\}$, where $k \in [N]$. If $k=1$, we get $\mathcal{Y} = \{\bm{e}_1, ..., \bm{e}_N\}$ and $\mathrm{conv}(\mathcal{Y}) = \triangle_N$, recovering the  scenario studied in \S\ref{sec:probabilistic}. 
For larger $k$, $|\mathcal{Y}| = {N \choose k} \gg N$. 
With a simple rescaling, the resulting marginal polytope is equivalent to the capped probability simplex  described by \citet[\S 7.3]{blondel2020learning}. 
\end{example}

Given unary scores $\bm{\theta} \in \mathbb{R}^N$, 
the structure with the largest score 
is 
$\bm{y}^* = \argmax_{\bm{y} \in \mathcal{Y}} \bm{\theta}^\top \bm{y} = \argmax_{\bm{y} \in \mathrm{conv}(\mathcal{Y})} \bm{\theta}^\top \bm{y}$, where the last equality comes from the fact that $\mathrm{conv}(\mathcal{Y})$ is a polytope, therefore the maximum is attained at a vertex. 
As in \eqref{eq:rpm}, we consider a regularized prediction version of this problem via a convex regularizer $\Omega : \mathrm{conv}(\mathcal{Y}) \rightarrow \mathbb{R}$:
\begin{align}\label{eq:sparsemap}
\hat{\bm{y}}_\Omega(\bm{\theta}) := \argmax_{\bm{y} \in \mathrm{conv}(\mathcal{Y})} \bm{\theta}^\top \bm{y} - \Omega(\bm{y}). 
\end{align}
By choosing $\Omega(\bm{y}) = \frac{1}{2}\|\bm{y}\|^2 + I_{\mathrm{conv}(\mathcal{Y})}(\bm{y})$, 
we obtain \textbf{SparseMAP}, a structured version of sparsemax which
can be computed efficiently via an active set algorithm as long as an argmax oracle is available
\citep{niculae2018sparsemap}.

\subsection{General case: factor graph, high order interactions}\label{sec:structured_high_order}

In general, we may want to consider soft interactions among patterns, for example due to temporal dependencies, hierarchical  structure, etc. 
These interactions can be expressed as a bipartite \textbf{factor graph} $(V, F)$, where $V = \{1, ..., N\}$ are variable nodes (associated with the patterns) and $F \subseteq 2^V$ are factor nodes representing the interactions \citep{kschischang2001factor}. 
A structure can be represented as a bit vector $\bm{y} = [\bm{y}_V; \bm{y}_F]$, where $\bm{y}_V$ and $\bm{y}_F$ indicate configurations of variable and factor nodes, respectively. 
The SparseMAP transformation has the same form \eqref{eq:sparsemap} with $\Omega(\bm{y}) = \frac{1}{2} \|\bm{y}_V\|^2 + I_{\mathrm{conv}(\mathcal{Y})}(\bm{y})$ (note that only the unary variables are regularized). 
Full details are given in App.~\ref{sec:factor_graphs}. 

\begin{example}[sequential $k$-subsets]\label{ex:seqksubsets} 
Consider the $k$-subset problem of Example~\ref{ex:ksubsets} but now with a sequential structure. 
This can be represented as a pairwise factor graph $(V, F)$ where $V = \{1, ..., N\}$ and $F = \{(i, i+1)\}_{i=1}^{N-1}$. 
The budget constraint forces exactly $k$ of the $N$ variable nodes to have the value $1$. 
The set $\mathcal{Y}$ contains all bit vectors satisfying these constraints as well as consistency among the variable and factor assignments. 
To promote consecutive memory items to be both retrieved or neither retrieved, one can define attractive Ising higher-order (pairwise) scores $\bm{\theta}_{(i, i+1)}$, 
in addition to the unary scores. 
The MAP inference problem can be solved with dynamic programming in runtime $\mathcal{O}(Nk)$, and the SparseMAP transformation can be computed with the active set algorithm \citep{niculae2018sparsemap} by iteratively calling this MAP oracle. 
\end{example}

\subsection{Structured Fenchel-Young losses and margins}


The notion of margin in Def.~\ref{def:margin} can  be extended to the structured case \citep[Def.~5]{blondel2020learning}:

\begin{definition}[Structured margin]
A loss $L(\bm{\theta}; \bm{y})$ has a \textbf{structured margin} if $\exists 
0 \le  m < \infty
$ 
such that   
$\forall \bm{y} \in \mathcal{Y}$:
\begin{align*}
\bm{\theta}^\top \bm{y} \ge \max_{\bm{y}' \in \mathcal{Y}} \left( \bm{\theta}^\top \bm{y}' + \frac{m}{2}\|\bm{y} - \bm{y}' \|^2 \right) \,\, \Rightarrow \,\, L(\bm{\theta}; \bm{y}) = 0.
\end{align*} 
The smallest such $m$ is called the margin of $L$.
\end{definition}

Note that this generalizes the notion of margin in Def.~\ref{def:margin}, recovered when $\mathcal{Y} = \{\bm{e}_1, ..., \bm{e}_N\}$. Note also that, since we are assuming  $\mathcal{Y} \subseteq \{0, 1\}^L$, the term $\|\bm{y} - \bm{y}' \|^2$ is a Hamming distance, which counts how many bits need to be flipped to transform $\bm{y}'$ into $\bm{y}$. A well-known example of a loss with a structured separation margin is the structured hinge loss \citep{taskar2003max,tsochantaridis2005large}. 

We show below that the SparseMAP loss has a structured margin (our result, proved in App.~\ref{sec:proof_prop_sparsemap_margin}, extends that of \citet{blondel2020learning}, who have shown this only for structures without high order interactions):

\begin{figure*}[t]
  \centering
  \includegraphics[width=0.49\textwidth]{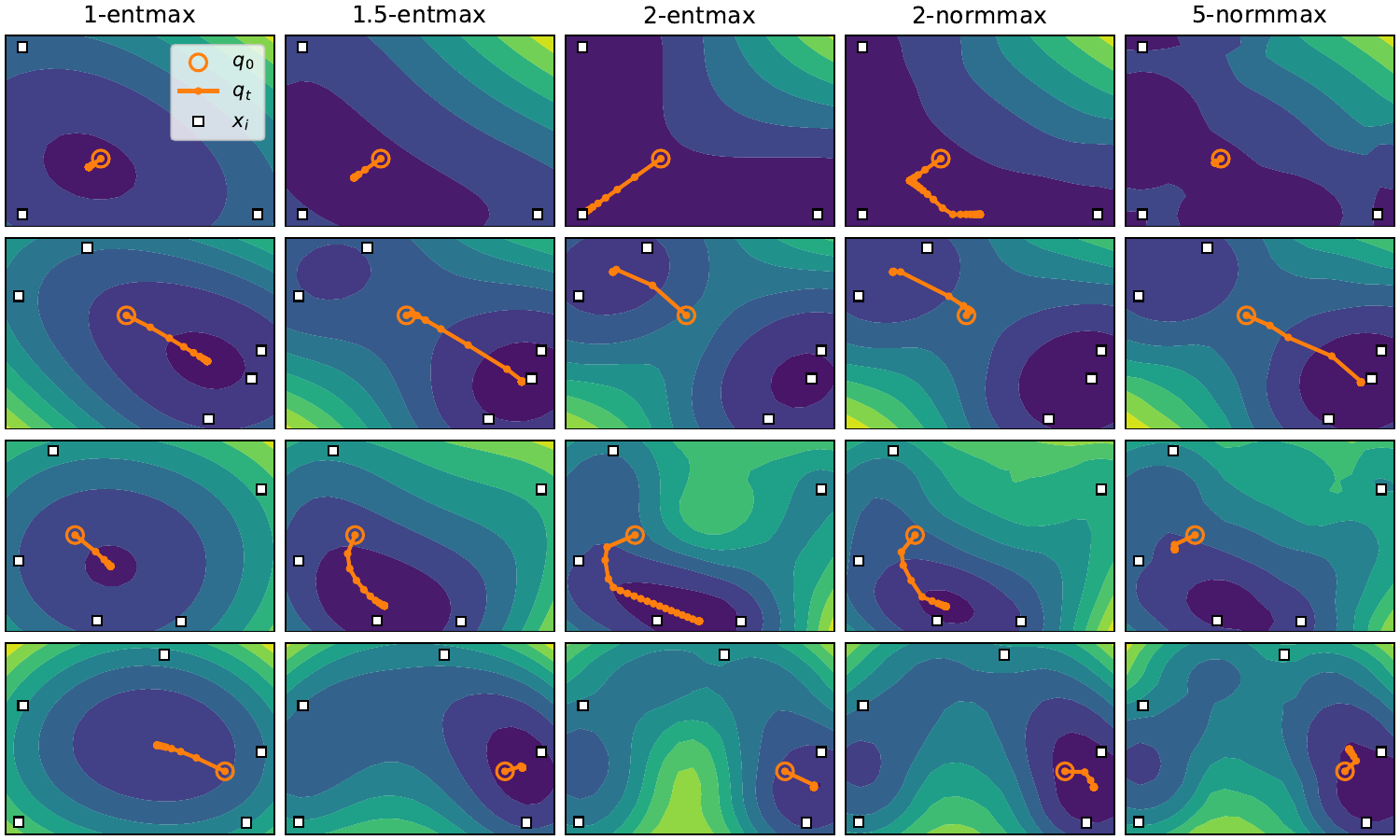}\hspace{2pt}%
  \includegraphics[width=0.49\textwidth]{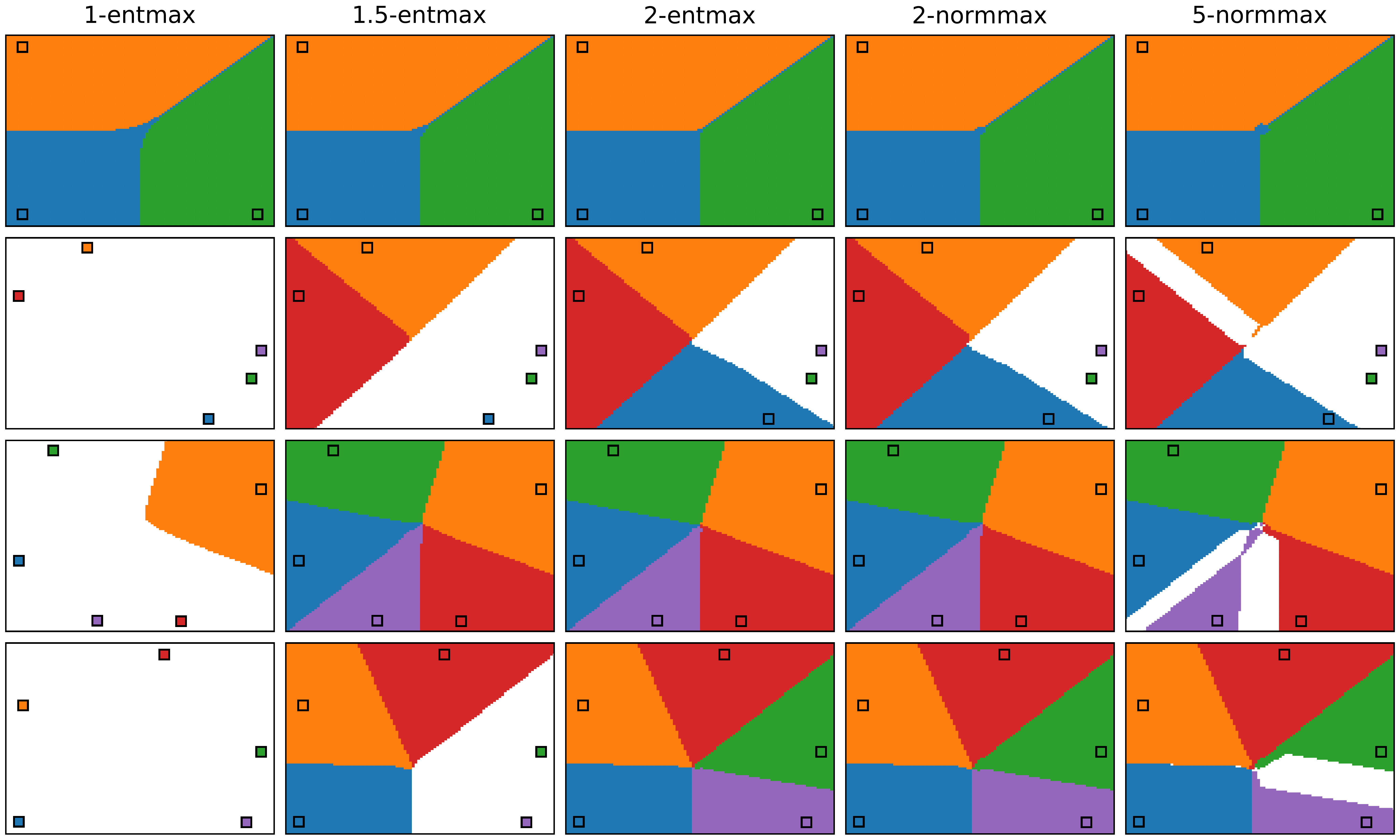}
  \caption{Left: contours of the energy function and optimization trajectory of the CCCP iteration ($\beta = 1$). Right: attraction basins associated with each pattern. (White sections do not converge to a single pattern but to a metastable state; $\beta = 10$ (a larger $\beta$ is needed to allow for the $1$-entmax to get $\epsilon$-close to a single pattern); for $\alpha = 1$ we allow a tolerance of $\epsilon = .01$). Additional plots for different $\beta$ can be found in App. \ref{sec:HDBA}.}
  \label{fig:overall}
\end{figure*}

\begin{proposition}\label{prop:sparsemap_margin}
Let $\mathcal{Y} \subseteq \{0, 1\}^L$ be contained in a sphere, i.e., for some $r>0$, $\|\bm{y}\| = r$ for all $\bm{y} \in \mathcal{Y}$. 
Then:
\begin{enumerate}
\item If there are no high order interactions, then the SparseMAP loss has a structured margin $m=1$. 
\item If there are high order interactions and, for some $r_V$ and $r_F$ with $r_V^2 + r_F^2 = r^2$, we have $\|\bm{y}_V\| = r_V$ and $\|\bm{y}_F\| = r_F$ for any $\bm{y} = [\bm{y}_V; \bm{y}_F]\in \mathcal{Y}$, then the SparseMAP loss has a structured margin $m \le 1$.
\end{enumerate}
\end{proposition}
The assumptions above are automatically satisfied with the factor graph construction in \S\ref{sec:structured_high_order}, with $r_V^2 = |V|$, $r_F^2 = |F|$, and $r^2 = |V| + |F|$. 
For the $k$-subsets example, we have $r^2 = k$, and for the sequential $k$-subsets example, we have $r_V^2 = N$, $r_F^2 = N-1$, and $r^2 = 2N-1$.

\subsection{Guarantees for retrieval of pattern associations}


We now consider a structured HFY network using SparseMAP. We obtain the following updates:
\begin{align}\label{eq:sparsemap_hfy_update_rule}
    \bm{q}^{(t+1)} = \bm{X}^\top \mathrm{SparseMAP}(\beta \bm{X}\bm{q}^{(t)}).
\end{align} 
%
%
%
In the structured case, we aim to  retrieve not individual patterns but pattern associations of the form $\bm{X}^\top \bm{y}$, where $\bm{y} \in \mathcal{Y}$. Naturally, when $\mathcal{Y} = \{\bm{e}_1, ..., \bm{e}_N\}$, we recover the usual patterns, since $\bm{x}_i = \bm{X}^\top \bm{e}_i$. 
We define the separation of pattern association $\bm{y}_i \in \mathcal{Y}$ from data as $\Delta_i = \bm{y}_i^\top \bm{X} \bm{X}^\top \bm{y}_i - \max_{j \ne i} \bm{y}_i^\top \bm{X} \bm{X}^\top \bm{y}_j$. 
The next proposition, proved in App.~\ref{sec:proof_prop_stationary_single_iteration_sparsemap}, states conditions for exact convergence in a single iteration, generalizing Prop.~\ref{prop:separation}. 

\begin{proposition}[Exact structured retrieval]\label{prop:stationary_single_iteration_sparsemap}
Let $\Omega(\bm{y})$ be the SparseMAP regularizer and assume the conditions of Prop.~\ref{prop:sparsemap_margin} hold. 
Let $\bm{y}_i \in \mathcal{Y}$ be such that  
$\Delta_i \ge \frac{D_i^2}{2\beta}$, where $D_i = \max \|\bm{y}_i - \bm{y}_j\| \le 2r$. Then, $\bm{X}^\top \bm{y}_i$ is a stationary point of the Hopfield energy. 
In addition, if $\bm{q}^{(0)}$ satisfies ${\bm{q}^{(0)}} ^\top \bm{X}^\top (\bm{y}_i - \bm{y}_j) \ge\frac{D_i^2}{2\beta}$ for all $j\ne i$, then the update rule 
\eqref{eq:sparsemap_hfy_update_rule} converges to the pattern association $\bm{X}^\top \bm{y}_i$ in one iteration. 
Moreover, 
if $$\Delta_i \ge \frac{D_i^2}{2\beta} + \epsilon \min \{\sigma_{\max}(\bm{X})D_i, MD_i^2\},$$   
where $\sigma_{\max}(\bm{X})$ is the spectral norm of $\bm{X}$ and $M = \max_k \|\bm{x}_k\|$, then any $\bm{q}^{(0)}$ $\epsilon$-close to  $\bm{X}^\top \bm{y}_i$  will converge to $\bm{X}^\top \bm{y}_i$ in one iteration.
\end{proposition}

Note that the bound above on $\Delta_i$ includes as a particular case the unstructured bound in Prop.~\ref{prop:separation} applied to sparsemax (entmax with $\alpha=2$, which has margin $m = 1/(\alpha-1) = 1$), since for $\mathcal{Y} = \triangle_N$ we have $r = 1$ and $D_i = \sqrt{2}$, which leads to the condition $\Delta_i \ge \beta^{-1} + 2M\epsilon$. 

For the particular case of the $k$-subsets problem (Example~\ref{ex:ksubsets}), we have $r = \sqrt{k}$ and $D_i=\sqrt{2k}$, leading to the condition $\Delta_i  \ge \frac{k}{\beta} + 2Mk\epsilon$. This recovers sparsemax when $k=1$. 

For the sequential $k$-subsets problem in Example~\ref{ex:seqksubsets}, we have $r = 2N-1$. Noting that any two distinct $\bm{y}$ and $\bm{y}'$ differ in at most $2k$ variable nodes, and since each variable node can affect 6 bits (2 for $\bm{y}_V$ and 4 for $\bm{y}_F$), the Hamming distance between  $\bm{y}$ and $\bm{y}'$ is at most $12k$, therefore we have $D_i = \sqrt{12k}$, which leads to the condition $\Delta_i  \ge \frac{6k}{\beta} + 12Mk\epsilon$.

\section{Experiments}\label{sec:experiments}

We now present experiments using synthetic and real-world datasets to validate our  theoretical findings  and illustrate the usefulness of sparse and structured Hopfield networks.

\subsection{Hopfield dynamics and basins of attraction}

Fig.~\ref{fig:overall} shows optimization trajectories and basins of attraction for various queries and artificially generated pattern configurations for the two families of sparse transformations, $\alpha$-entmax and $\alpha$-normmax. We use $\alpha \in \{1, 1.5, 2\}$ for $\alpha$-entmax and $\alpha \in \{2, 5\}$ for $\alpha$-normmax (where we apply the bisection algorithm described in App.~\ref{sec:normmax}). As $\alpha$ increases, $\alpha$-entmax converges more often to a single pattern, whereas $\alpha$-normmax tends to converge towards an attractor which is a uniform average of some patterns.

\subsection{Metastable state distributions in MNIST}
\begin{table*}[t]
\small
    \centering
    \caption{Distribution of metastable state 
    (in $\%$) in MNIST. The 
    training set 
    is memorized and the 
    test set is used as queries.}
    \label{tab:metastable}
    \vspace{-0.1cm}
    \begin{tabular}{c|S[table-format=2.1] S[table-format=1.1] S[table-format=1.1] S[table-format=1.1] S[table-format=1.1] S[table-format=1.1]  S[table-format=2.1]  S[table-format=1.1]| S[table-format=1.1] S[table-format=1.1] S[table-format=1.1] S[table-format=1.1] S[table-format=2.1] S[table-format=1.1] S[table-format=1.1] S[table-format=1.1] S[table-format=1.1] S[table-format=1.1]}
    \toprule
    Metastable & \multicolumn{8}{c}{$\beta = 0.1$} & \multicolumn{8}{c}{$\beta = 1$} \\
     State& \multicolumn{3}{c}{$\alpha$-entmax} & \multicolumn{2}{c}{$\alpha$-normmax} & \multicolumn{3}{c|}{$k$-subsets} & \multicolumn{3}{c}{$\alpha$-entmax} & \multicolumn{2}{c}{$\alpha$-normmax} & \multicolumn{3}{c}{$k$-subsets} \\
     Size & {1} & {1.5} & \multicolumn{1}{c|}{2} & {2} & \multicolumn{1}{c|}{5} & {2} & {4} & \multicolumn{1}{c|}{8} & {1} & {1.5} & \multicolumn{1}{c|}{2} & {2} & \multicolumn{1}{c|}{5} & {2} & {4} & {8} \\
    \midrule
        
    1 & 3.5 & 69.2 & 88.1 & 81.4 & 51.4& 0.0 & 0.0 & 0.0 & 97.8 & 99.9 & 100.0 & 100.0 &99.8 & 0.0 & 0.0 & 0.0 \\
    2 & 2.1 & 8.6 & 5.2 & 6.7 & 31.4& 87.3 & 0.0 & 0.0 & 0.9 & 0.1 & 0.0 & 0.0 & 0.2&99.9 & 0.0 & 0.0 \\
    3 & 1.6 & 3.9 & 2.6 & 1.9 & 7.0& 6.1 & 0.0 & 0.0 & 0.4 & 0.0 & 0.0 & 0.0 & 0.0&0.1 & 0.0 & 0.0 \\
    4 & 1.2 & 2.3 & 1.6 & 1.0 & 2.1& 2.5 & 80.0 & 0.0 & 0.3 & 0.0 & 0.0 & 0.0 &0.0 &0.0 & 99.3 & 0.0 \\
    5 & 1.2 & 1.6 & 1.1 & 0.9 &1.5& 2.0 & 11.9 & 0.0 & 0.2 & 0.0 & 0.0 & 0.0 & 0.0&0.0 & 0.7 & 0.0 \\
    6 & 0.9 & 0.9 & 0.8 & 0.5 &1.5& 1.1 & 4.4 & 0.0 & 0.1 & 0.0 & 0.0 & 0.0 & 0.0& 0.0&  0.1 & 0.0 \\
    7 & 1.1 & 0.6 & 0.4 & 0.4 & 1.3&0.6 & 2.1 & 0.0 & 0.1 & 0.0 & 0.0 & 0.0 &0.0 & 0.0 & 0.0 & 0.0 \\
    8 & 0.8 & 0.6 & 0.1 & 0.8 & 1.0&0.2 & 1.0 & 60.0 & 0.1 & 0.0 & 0.0 & 0.0 & 0.0& 0.0 & 0.0 & 95.0 \\
    9 & 1.0 & 0.3 & 0.0 & 0.5 &0.8& 0.1 & 0.4 & 26.0 & 0.1 & 0.0 & 0.0 & 0.0 & 0.0& 0.0 & 0.0 & 4.7 \\
    10 & 1.1 & 0.1 & 0.0 & 0.5 &0.6& 0.0 & 0.1 & 9.2 & 0.1 & 0.0 & 0.0 & 0.0 & 0.0& 0.0 & 0.0 & 0.2 \\
    10$^+$ & 85.5 & 11.9 & 0.1 & 5.4 & 1.4 & 0.1 & 0.0 & 4.8 & 0.1 & 0.0 & 0.0 & 0.0 &0.0 & 0.0 & 0.0 & 0.1 \\
        \bottomrule
    \end{tabular}
\end{table*}

\begin{table*}[t]
    \caption{Results for MIL. We show accuracies for MNIST and ROC AUC for MIL benchmarks, averaged across 5 runs.}
    \vspace{-0.1cm}
    \centering
    \small
    \begin{tabular}{lcccccccccccccccc}
        \toprule
        \multicolumn{1}{c}{} & \multicolumn{4}{c}{MNIST} & \multicolumn{3}{c}{MIL benchmarks} \\
        \cmidrule(lr){2-5}
        \cmidrule(lr){6-8}
        Methods & $K$=1 & $K$=2 & $K$=3 & $K$=5 & Fox & Tiger & Elephant  \\
        \midrule
        1-entmax (softmax) & $\mathbf{98.4\pm0.2}$ & $94.6\pm0.5$ & $91.1\pm0.5$ & $89.0\pm0.3$ & $66.4\pm2.0$ & $87.1\pm1.6$ & $92.6\pm0.6$\\
        1.5-entmax & $97.6\pm0.8$&$96.0\pm 0.9$ & $90.4\pm1.1$& $92.4\pm1.4$& $66.3\pm2.0$ & $87.3\pm1.5$ & $92.4\pm1.0$ \\
        2.0-entmax (sparsemax) &$97.9\pm0.2$& $96.7\pm0.5$ & $92.9\pm0.9$& $91.6\pm1.0$ & $66.1\pm0.6$ & $\mathbf{87.7\pm1.4}$ & $91.8\pm0.6$\\
        2.0-normmax &$97.9\pm0.3$& $96.6\pm0.6$& $93.9\pm0.7$& $92.4\pm0.7$& $66.1\pm2.5$ & 
        $86.4\pm0.8$ & $92.4\pm0.7$\\
        5.0-normmax &$98.2\pm0.5$& $97.2\pm0.3$ & $95.8\pm0.4$& $93.2\pm0.5$ & $66.4\pm2.3$& $85.5\pm0.6$ & $93.0\pm0.7$
        \\
        SparseMAP, $k=2$ & $97.9\pm0.3$ & $\mathbf{97.7\pm0.3}$ & $95.1\pm0.5$ & $92.6\pm1.1$ & $66.8\pm2.7$ & $85.3\pm0.5$ & $\mathbf{93.2\pm0.7}$ \\
        SparseMAP, $k=3$ &$98.0\pm0.6$ & $96.1\pm1.0$ & $\mathbf{96.5\pm0.5}$& $92.2\pm1.2$& $\mathbf{67.4\pm2.0}$ & $86.1\pm0.8$ & $92.6\pm1.7$\\
        SparseMAP, $k=5$ &$98.2\pm0.4$&$96.2\pm1.4$& $95.1\pm1.1$ & $\mathbf{95.1\pm1.5}$& $67.0\pm2.0$ & $86.3\pm0.8$ & $91.2\pm1.0$\\
        \bottomrule
    \end{tabular}
    \label{tab:MIL}
\end{table*}

We next investigate how often our Hopfield networks converge to metastable states, a crucial aspect for understanding the network's dynamics. To elucidate this, we examine  $\hat{\bm{y}}_\Omega(\beta\bm{X}\bm{q}^{(t)})$ for the MNIST dataset, probing the number of nonzeros of these vectors. We set a threshold $>0.01$ for the softmax method (1-entmax). For the sparse transformations we do not need a threshold, since they have exact retrieval. 

Results in Tab.~\ref{tab:metastable} suggest that $\alpha$-entmax is capable of retrieving single patterns for higher values of $\alpha$. Despite $\alpha$-normmax's ability to induce sparsity, we observe that it tends to stabilize 
in
small but persistent metastable states as $\alpha$ is increased, whereas
SparseMAP with $k$-subsets 
is capable
of retrieving associations of $k$ patterns, as expected.

\subsection{Multiple instance learning}

In multiple instance learning (MIL), instances are grouped into ``bags'' and a bag is labeled as positive if it contains at least one instance from a given class. 
We also consider a extended variant, denoted $K$-MIL, where bags are considered positive if they contain $K$ or more positive instances. 

\citet{ramsauer2020hopfield} tackle MIL via a Hopfield pooling layer, where the query $\bm{q}$ is learned and the keys $\bm{X}$ are instance embeddings. 
We experiment with sparse Hopfield pooling layers using our proposed $\alpha$-entmax and $\alpha$-normmax transformations (\S\ref{sec:probabilistic}), as well as structured Hopfield pooling layers using SparseMAP with $k$-subsets (\S\ref{sec:structured}), varying $\alpha$ and $k$ in each case. 
Note that 1-entmax recovers \citet{ramsauer2020hopfield} and 2-entmax recovers \citet{hu2023sparse}. 
We run these models for $K$-MIL problems in the MNIST dataset (choosing `9' as target) 
and in three MIL benchmarks: Elephant, Fox, and Tiger \cite{ilse2018attention}.
Further details can be found in App. \ref{sec:MNIST_Experimental_Details} and \ref{sec:MIL_bench_details}.


Tab.~\ref{tab:MIL} shows the results. We observe that for MNIST, $K=1$, $1$-entmax surprisingly outperforms the remaining methods. Normmax shows consistent results across datasets achieving near optimal performance, arguably due to its ability to adapt to near-uniform metastable states of varying size. We also observe that, for $K>1$, the $k$-subsets approach achieves top performance when $k=K$, as expected. 
We also see that, in the MIL benchmarks, SparseMAP pooling surpasses sparse pooling variants for 2 out of 3 datasets. 

\begin{table*}[t]
\caption{Text rationalization results. We report mean and min/max $F_1$ scores across ﬁve random seeds on test sets for all datasets but Beer, where we report
MSE. HardKuma and SPECTRA results are taken from \cite{guerreiro2021spectra}. We also report human rationale overlap (HRO) as $F_1$ score. We bold the best-performing system(s).}
\centering
\small
\begin{tabular}{lcccccc}
\toprule
& SST$\uparrow$ & AgNews$\uparrow$ & IMDB$\uparrow$ & Beer (MSE) $\downarrow$  &Beer (HRO) $\uparrow$\\
\midrule
HardKuma  \citep{bastings2019interpretable} & .80 {\small(.80/.81)} & .90 {\small(.87/.88)} & .87 {\small(.90/.91)} & .019 {\small(.016/.020)} & .37 {\small(.00/.90)}  \\
SPECTRA \citep{guerreiro2021spectra} & .80 {\small(.79/.81)} & .92 {\small(.92/.93)} & \textbf{.90} {\small(.89/.90)} & \textbf{.017} {\small(.016/.019)} & .61 {\small(.56/.68)}  \\
\midrule
SparseMAP $k$-subsets (ours) & \textbf{.81} {\small(.81/.82)}  & \textbf{.93} {\small(.92/.93)} & \textbf{.90} {\small(.90/.90)} & \textbf{.017} {\small(.017/.018)} & .42 {\small(.29/.62)}   \\
SparseMAP seq. $k$-subsets (ours) & \textbf{.81} {\small(.80/.83)}& \textbf{.93} {\small(.93/.93)} & \textbf{.90} {\small(.90/.90)} & .020 {\small(.018/.021)} & \textbf{.63} {\small(.49/.70)} \\
\bottomrule
\end{tabular}
\label{tab:spectra}
\end{table*}

\subsection{Structured Rationalizers}

\begin{figure}[t]
    \centering
    \includegraphics[width=0.95\columnwidth]{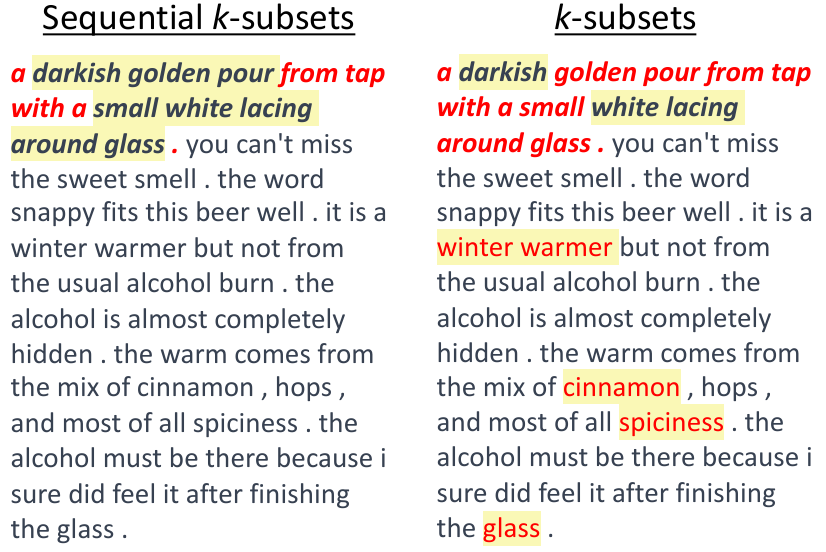}
    \caption{Example of human rationale overlap for the aspect ``appearance''. The \hl{yellow highlight} indicates the model's rationale, while \textbf{\textit{italicized and bold font}} represents the human rationale. \textcolor{red}{Red font} identifies mismatches with human annotations. SparseMAP  with sequential $k$-subsets prefers more contiguous  rationales, which better match humans. Additional examples are shown in App. \ref{sec:text_rationalization_details}.}
    \label{fig:overlap}
    \vspace{-0.2cm}
\end{figure}

We experiment with rationalizer models in sentiment prediction tasks, where the inputs are sentences or documents in natural language and the rationales are text highlights (see Fig.~\ref{fig:overlap}). 
These models, sometimes referred as select-predict or explain-predict models \citep{jacovi2021aligning,zhang2021explain}, consist of a rationale generator and a predictor. The generator processes the input text and extracts the rationale as a subset of words to be highlighted, and the predictor classifies the input based solely on the extracted rationale, which generally involves concealing non-rationale words through the application of a binary mask. 
Rationalizers are usually trained end-to-end, and the discreteness of the latent rationales is either handled with stochastic methods via score function estimators or the reparametrization trick \citep{lei2016rationalizing,bastings2019interpretable}, or with deterministic methods via structured continuous relaxations \citep{guerreiro2021spectra}. In either case, the model imposes sparsity and contiguity penalties to ensure rationales are short and tend to extract  adjacent words.  
Our model architecture is adapted from SPECTRA \citep{guerreiro2021spectra}, but the combination of the generator and predictor departs from prior approaches \citep{lei2016rationalizing,bastings2019interpretable,guerreiro2021spectra} in which the predictor does not ``mask'' the input tokens; instead, it takes as input the pooled vector that results from the Hopfield pooling layer (either a sequential or non-sequential SparseMAP $k$-subsets layer). 
By integrating this Hopfield pooling layer into the predictor, we transform the sequence of word embeddings into a single vector from which the prediction is made. The rationale is formed by the pattern associations (word tokens) extracted by the Hopfield layer. 

Tab.~\ref{tab:spectra} shows the results in the downstream task (classification for SST, AgNews, IMDB; regression for BeerAdvocate) as well as the $F_1$ overlap with human rationales for the BeerAdvocate dataset \citep{mcauley2012learning}. 
Compared to strong baselines \citep{bastings2019interpretable,guerreiro2021spectra}, our proposed methods achieve equal or slightly superior performance for all datasets. Moreover, our sequential $k$-subsets model outperforms the baselines in terms of overlap with human rationales. This is explained by the fact that human rationales tend to contain adjacent words, which is encouraged by our sequential model. 
Additional details and results with other baselines are shown in App.~\ref{sec:text_rationalization_details}.

\section{Related Work}

Recent research on modern Hopfield networks and dense associative memories includes \citep[\textit{inter alia}]{krotov2016dense,demircigil2017model,ramsauer2020hopfield,millidge2022universal,hoover2023energy}. 
The closest to our work is \citet{hu2023sparse}, who proposed {sparse Hopfield networks} (equivalent to our 2-entmax, i.e., sparsemax) 
and derived retrieval error bounds tighter than the dense analog. 
Concurrently to our work, \citet{wu2023stanhop} further proposed a ``generalized sparse Hopfield model'', which corresponds to $\alpha$-entmax with learnable $\alpha$, and applied it successfully to time series prediction problems. 
However, neither \citet{hu2023sparse} or \citet{wu2023stanhop} seem to have realized the possibility of \textbf{exact} retrieval enabled by sparse transformations. 
Our paper fills this gap by providing a unified framework for sparse Hopfield networks with stronger theoretical guarantees for retrieval and coverage (cf. Prop.~\ref{prop:separation}--\ref{prop:storage}).  
Our framework generalizes their constructions and widens the scope to new families, such as $\alpha$-normmax, for which we provide an effective bisection algorithm (Alg.~\ref{alg:normmax_bisection} in App.~\ref{sec:normmax}). 

The link with {Fenchel-Young losses} \citep{blondel2020learning} is a key dimension of our work. Many results derived therein, such as the margin conditions, were found to have direct application to sparse Hopfield networks. In our paper, we extend some of their results, such as the structured margin of SparseMAP (Prop.~\ref{prop:sparsemap_margin}), key to establish exact retrieval of pattern associations (Prop.~\ref{prop:stationary_single_iteration_sparsemap}). 
Our $k$-subsets example relates to the top-$k$ retrieval model of  \citet{davydov2023retrieving}: their model differs from ours as it uses an entropic regularizer not amenable to sparsity, making exact retrieval impossible. 


Our sparse and structured Hopfield layers with sparsemax and SparseMAP involve a quadratic regularizer, which relates to the differentiable layers of \citet{amos2017optnet}. The use of SparseMAP and its active set algorithm \citep{niculae2018sparsemap} allows to exploit the structure of the problem to ensure efficient Hopfield updates and implicit derivatives.

\section{Conclusions}

We presented a unified framework for sparse and structured Hopfield networks. 
Our framework hinges on a broad family of energy functions
written as a difference of a quadratic regularizer and a Fenchel-Young loss, parametrized by
a generalized negentropy function. 
A core result of our paper is the link between the margin property of certain Fenchel-Young losses and sparse Hopfield networks with provable conditions for exact retrieval. 
We further extended 
this framework to incorporate structure via the SparseMAP transformation, which is able to
retrieve pattern associations instead of a single
pattern. 
Empirical evaluation confirmed the usefulness of our approach in multiple
instance learning and text rationalization problems.

\section*{Impact Statement}

Hopfield networks are increasingly relevant for practical applications and not
only as theoretical models. While our work is mostly a theoretical advancement,
promising experimental results signal potentially wider impact. Sparse HFY 
networks are applicable in the same scenarios where modern Hopfield networks
would be, and we do not foresee any specific societal consequences of sparse 
transformations or exact retrieval in such cases. In the structured case,
the practicioner has more hands-on control for encoding inductive biases
through the design of a factor graph and the choice of its parameters
(e.g., $k$). These choices may in turn reflect human biases and 
have societal implications: for example, contiguous rationales might be well 
suited for English but not for other languages. We encourage care when designing
such methods for practical applications.

\section*{Acknowledgments} 

This work was supported by the European Research Council (DECOLLAGE, ERC-2022-CoG 101088763), by the Portuguese Recovery and Resilience Plan through project C645008882-00000055 (Center for Responsible AI), by Fundação para a Ciência e Tecnologia through contract UIDB/50008/2020,
and by the Dutch Research Council (NWO) via VI.Veni.212.228.


\bibliography{bib}
\bibliographystyle{icml2024}

\newpage

\appendix
\onecolumn

\section{Generalized Negentropies}\label{sec:generalized_negent}

We recall here the definition of generalized negenetropies from \citet[\S4.1]{blondel2020learning}:

\begin{definition}
\label{def:generalized_negent}
   A function \(\Omega : \triangle_{N} \to \mathbb{R}\)
   is a generalized negentropy iff it satisfies the following conditions:
   \begin{enumerate}
      \item Zero negentropy: $\Omega(\bm{y})=0$
      if \(\bm{y}\) is a one-hot vector (delta distribution), i.e.,
      \(\bm{y}=\bm{e}_i\) for any \(i \in \{1,\ldots,N\}\).
      \item Strict convexity:
      \(\Omega\left((1-\lambda)\bm{y} + \lambda \bm{y}'\right)
      < (1-\lambda)\Omega(\bm{y}) + \lambda \Omega(\bm{y}')
      \) for $\lambda \in \,\, ]0,1[$ and $\bm{y}, \bm{y}' \in \triangle_N$ with $\bm{y} \ne \bm{y}'$.  
      \item Permutation invariance:
      \(\Omega(\bm{Py})=\Omega(\bm{y})\) for any permutation matrix \(\bm{P}\)
      (i.e., square matrices with a single 1 in each row and each column, zero elsewhere).
   \end{enumerate}
\end{definition}

This definition implies that $\Omega \le 0$ and that $\Omega$ is minimized when $\bm{y} = \mathbf{1}/N$ is the uniform distribution \citep[Prop.~4]{blondel2020learning}. This justifies the name ``generalized negentropies.''

\section{Bisection Algorithm for the Normmax Transformation}\label{sec:normmax}

We derive here expressions for the normmax transformation along with a bisection algorithm to compute this transformation for general $\alpha$. 

Letting $\Omega(\bm{y}) = -1 + \|\bm{y}\|_\alpha + I_{\triangle_N}(\bm{y})$ be the norm entropy, we have 
\begin{align}
    (\nabla \Omega^*)(\bm{\theta}) = \arg\max_{\bm{y} \in \triangle_N} \bm{\theta}^\top \bm{y} - \|\bm{y}\|_q.
\end{align}
The Lagrangian function is $L(\bm{y}, \bm{\lambda}, \mu) = -\bm{\theta}^\top \bm{y} + \|\bm{y}\|_\alpha - \bm{\lambda}^\top\bm{y} + \mu (\mathbf{1}^\top \bm{y} - 1)$. 
Equating the gradient to zero and using the fact that $\nabla \|\bm{y}\|_\alpha = \left(\frac{\bm{y}}{\|\bm{y}\|_\alpha}\right)^{\alpha-1}$, we get:
\begin{align}\label{eq:kkt}
    \mathbf{0} = \nabla_{\bm{y}} L(\bm{y}, \bm{\lambda}, \mu) = 
    -\bm{\theta} + \left(\frac{\bm{y}}{\|\bm{y}\|_\alpha}\right)^{\alpha-1} - \bm{\lambda} + \mu \mathbf{1}.
\end{align}
The complementarity slackness condition implies that, if $y_i > 0$, we must have $\lambda_i = 0$, therefore, we have for such $i \in \mathrm{supp}(\bm{y})$: 
\begin{align}\label{eq:kkt_mu}
    -\theta_i + \left(\frac{y_i}{\|\bm{y}\|_\alpha}\right)^{\alpha-1} + \mu = 0 \quad \Rightarrow \quad y_i = (\theta_i - \mu)^{\frac{1}{\alpha-1}} \|\bm{y}\|_\alpha.
\end{align}
Since we must have $\sum_{i \in \mathrm{supp}(\bm{y})} y_i = 1$, we obtain:
\begin{align}
    1 = \sum_{i \in \mathrm{supp}(\bm{y})} (\theta_i - \mu)^{\frac{1}{\alpha-1}} \|\bm{y}\|_\alpha \quad \Longrightarrow \quad \|\bm{y}\|_\alpha = \frac{1}{\sum_{i \in \mathrm{supp}(\bm{y})} (\theta_i - \mu)^{\frac{1}{\alpha-1}}}.
\end{align}
Combining the two last equations and noting that, from \eqref{eq:kkt}, we have $\theta_i < \mu_i$ for $i \notin \mathrm{supp}(\bm{y})$, we get, for $i \in [N]$:
\begin{align}\label{eq:normmax_solution}
y_i = \frac{(\theta_i - \mu)_+^{\frac{1}{\alpha-1}}}{\sum_{j \in \mathrm{supp}(\bm{y})} (\theta_j - \mu)_+^{\frac{1}{\alpha-1}}}.
\end{align}
Moreover, since $\sum_{i \in \mathrm{supp}(\bm{y})} y_i^\alpha = \|\bm{y}\|_\alpha^\alpha$, we obtain from \eqref{eq:kkt_mu}:
\begin{align}\label{eq:normmax_normalization_condition}
    \|\bm{y}\|_\alpha^\alpha = \sum_{i \in \mathrm{supp}(\bm{y})} (\theta_i - \mu)^{\frac{\alpha}{\alpha-1}} \|\bm{y}\|_\alpha^\alpha \quad \Rightarrow \quad \sum_{i \in \mathrm{supp}(\bm{y})} (\theta_i - \mu)^{\frac{\alpha}{\alpha-1}} = 1.
\end{align}

In order to compute the solution \eqref{eq:normmax_solution} we need to find $\mu$ satisfying \eqref{eq:normmax_normalization_condition}. This can be done with a simple bisection algorithm if we find a lower and upper bound on $\mu$. 

We have, from \eqref{eq:normmax_solution}, that 
$\mu = \theta_i - (y_i / \|\bm{y}\|_\alpha)^{\alpha-1}$ for any $i \in \mathrm{supp}(\bm{y})$. 
Letting $\theta_{\max} = \max_i \theta_i$ and $y_{\max} = \max_i y_i$, we have in particular that 
$\mu = \theta_{\max} - (y_{\max} / \|\bm{y}\|_\alpha)^{\alpha-1}$. 
We also have that $y_{\max} = \|\bm{y}\|_\infty \le \|\bm{y}\|_\alpha$, which implies $y_{\max} / \|\bm{y}\|_\alpha \le 1$. 
Since $1/N \le y_{\max} \le 1$ and $\|\bm{y}\|_\alpha \le 1$ for any $\bm{y} \in \triangle_N$,  
we also obtain 
$y_{\max} / \|\bm{y}\|_\alpha \ge (1/N) / 1 = N^{-1}$. 
Therefore we have
\begin{align}
\underbrace{\theta_{\max} - 1}_{\mu_{\min}} \, \le \, \mu \, \le \, \underbrace{\theta_{\max} - N^{1-\alpha}}_{\mu_{\max}}.
\end{align}
The resulting algorithm is shown as Alg.~\ref{alg:normmax_bisection}. 

\begin{algorithm}[t]
\caption{Compute $\alpha$-normmax by bisection.}\label{alg:normmax_bisection}
\begin{algorithmic}[1]
\STATE \textbf{Input:} Scores $\bm{\theta} = [\theta_1, ..., \theta_N]^\top \in \mathbb{R}^N$, parameter $\alpha > 1$.
\STATE \textbf{Output:} Probability vector $\bm{y} = [y_1, ..., y_N]^\top \in \triangle_N$. 
\STATE Define $\theta_{\max} \leftarrow \max_i \theta_i$
\STATE Compute $\mu_{\min} \leftarrow \theta_{\max} - 1$
\STATE Compute $\mu_{\max} \leftarrow \theta_{\max} - N^{1 - \alpha}$
\FOR{$t \in 1, \dots, T$}

\STATE  $\mu \leftarrow (\mu_{\min} + \mu_{\max}) / 2$
\STATE $Z \leftarrow \sum_j (\theta_j - \mu)_+^{\frac{\alpha}{\alpha-1}}$
\STATE \textbf{if} {$Z<1$}~\textbf{then}~$\mu_{\max}\leftarrow\mu$~%
\textbf{else}~$\mu_{\min}\leftarrow\mu$
\ENDFOR
\STATE Return $\bm{y} = [y_1, ..., y_N]^\top$ with 
$y_i = \frac{(\theta_i - \mu)_+^{\frac{1}{\alpha-1}}}{\sum_{j} (\theta_j - \mu)_+^{\frac{1}{\alpha-1}}}.$
\label{line:return_normalize}
\end{algorithmic}
\label{algo:bisect}
\end{algorithm}

\section{Structured Prediction with Factor Graphs}\label{sec:factor_graphs}

We define our notation and setup for structured prediction with factor graph representations, based on \citet{niculae2018sparsemap}. 

We assume the interactions among patterns (for example due to temporal dependencies, hierarchical or link structure, etc.) can be expressed as a bipartite \textbf{factor graph} $(V, F)$, where $V$ is a set of variable nodes and $F$ are factor nodes \citep{kschischang2001factor}. Each factor $f \in F$ is linked to a subset of variable nodes $V_f \subseteq V$. 
We assume each variable $v \in V$ can take one of $N_v$ possible values, and we denote by $\bm{y}_v \in \{0,1\}^{N_v}$ a one-hot vector indicating a value for this variable. 
Likewise, each factor $f \in F$ has $N_f$ possible configurations, with $N_f = \prod_{v \in V_f} N_v$, and we associate to it a one-hot vector $\bm{y}_f \in \{0,1\}^{N_f}$ indicating a configuration for that factor. 
The global configuration of the factor graph is expressed through the bit vectors $\bm{y}_V = [\bm{y}_v \,:\, v \in V] \in \{0, 1\}^{N_V}$ and $\bm{y}_F = [\bm{y}_f \,:\, f \in F] \in \{0, 1\}^{N_F}$, with $N_V = \sum_{v \in V} N_v$ and $N_F = \sum_{f \in F} N_f$. A particular structure is expressed through the bit vector $\bm{y} = [\bm{y}_V; \bm{y}_F] \in \{0, 1\}^{N_V + N_F}$. 
Finally, we define the set of \textbf{valid structures} $\mathcal{Y} \subseteq \{0, 1\}^{N_V + N_F}$ -- this set contains all the bit vectors which correspond to valid structures, which must satisfy consistency between variable and factor assignments, as well as any additional hard constraints. 

We associate \textbf{unary scores} $\bm{\theta}_V = [\bm{\theta}_v \,:\, v \in V] \in \mathbb{R}^{N_V}$ to  configurations of variable nodes and \textbf{higher-order scores} $\bm{\theta}_F = [\bm{\theta}_f \,:\, f \in F] \in \mathbb{R}^{N_F}$ to configurations of factor nodes. We denote $\bm{\theta} = [\bm{\theta}_V; \bm{\theta}_F] \in \mathbb{R}^{N_V + N_F}$. 
The problem of finding the highest-scoring structure, called the \textbf{maximum a posteriori} (MAP) inference problem, is 
\begin{align}\label{eq:map}
\bm{y}^* &= \argmax_{\bm{y} \in \mathcal{Y}} \bm{\theta}^\top \bm{y} = \argmax_{\bm{y} \in \mathrm{conv}(\mathcal{Y})} \bm{\theta}_V^\top \bm{y}_V + \bm{\theta}_F^\top \bm{y}_F.
\end{align}
As above, we consider regularized variants of \eqref{eq:map} via a convex regularizer $\Omega: \mathrm{conv}(\mathcal{Y}) \rightarrow \mathbb{R}$. 
\textbf{SparseMAP} corresponds to $\Omega(\bm{y}) = \frac{1}{2} \|\bm{y}_V\|^2$ (note that only the unary variables are regularized), which leads to the quadratic optimization problem \eqref{eq:sparsemap}.  \citet{niculae2018sparsemap} developed an effective and computationally efficient active set algorithm for solving this problem, which requires only a MAP oracle to solve instances of the problem \eqref{eq:map}. 


\section{Proofs of Main Text}

\subsection{Proof of Proposition~\ref{prop:bounds_cccp}}\label{sec:proof_prop_bounds_cccp}

We start by proving that $E(\bm{q}) \ge 0$.
    We show first that for any $\Omega$ satisfying conditions 1--3 above, we have
    \begin{align}\label{eq:upper_bound_fy}
    L_\Omega(\bm{\theta}; \mathbf{1}/N) \le \max_i \theta_i -\mathbf{1}^\top \bm{\theta}/N.
    \end{align}
    From the definition of $\Omega^*$ and the fact that $\Omega(\bm{y}) \ge \Omega(\mathbf{1}/N)$ for any $\bm{y} \in \triangle_{N}$,
    we have that, for any $\bm{\theta}$, $\Omega^*(\bm{\theta}) = \max_{\bm{y} \in \triangle_{N}} \bm{\theta}^\top \bm{y} - \Omega(\bm{y}) \le \max_{\bm{y} \in \triangle_{N}} \bm{\theta}^\top \bm{y} - \Omega(\mathbf{1}/N) = \max_i \theta_i - \Omega(\mathbf{1}/N)$,
    which leads to \eqref{eq:upper_bound_fy}.

    Let now $k = \arg\max_i \bm{q}^\top \bm{x}_i$, i.e., $\bm{x}_k$ is the pattern most similar to the query $\bm{q}$.
    We have
    \begin{align*}
        E(\bm{q}) &= -\beta^{-1} L_\Omega(\beta \bm{X} \bm{q}; \mathbf{1}/{N}) + \frac{1}{2} \|\bm{q} - \bm{\mu}_{\bm{X}}\|^2 + \frac{1}{2}(M^2 - \|\bm{\mu}_{\bm{X}}\|^2)\\
        &\ge
        -\beta^{-1} (\beta\max_i \bm{q}^\top \bm{x}_i - \beta \mathbf{1}^\top \bm{X}\bm{q}/N) + \frac{1}{2} \|\bm{q} - \bm{\mu}_{\bm{X}}\|^2 + \frac{1}{2}(M^2 - \|\bm{\mu}_{\bm{X}}\|^2)\\
        &= -\bm{q}^\top \bm{x}_k + \bm{q}^\top \bm{\mu}_{\bm{X}} + \frac{1}{2} \|\bm{q} - \bm{\mu}_{\bm{X}}\|^2 + \frac{1}{2}(M^2 - \|\bm{\mu}_{\bm{X}}\|^2)\\
        &= -\bm{q}^\top \bm{x}_k + \frac{1}{2} \|\bm{q}\|^2 + \frac{1}{2}M^2 \\
        &\ge -\bm{q}^\top \bm{x}_k + \frac{1}{2} \|\bm{q}\|^2 + \frac{1}{2}\|\bm{x}_k\|^2 \\
        &= \frac{1}{2}\|\bm{x}_k - \bm{q}\|^2 \ge 0.
    \end{align*}
    The zero value of energy is attained when $\bm{X} = \mathbf{1} \bm{q}^\top$ (all patterns are equal to the query), in which case $\bm{\mu}_{\bm{X}} = \bm{q}$, $M = \|\bm{q}\| = \|\bm{\mu}_{\bm{X}}\|$,  and we get $E_{\mathrm{convex}}(\bm{q}) = E_{\mathrm{concave}}(\bm{q}) = 0$.

    Now we prove the two upper bounds.
    For that, note that, for any $\bm{y} \in \triangle_{N}$, we have $0 \le L_{\Omega}(\bm{\theta}, \bm{y}) = L_{\Omega}(\bm{\theta}, \mathbf{1}/N) - \Omega(\mathbf{1}/N) + \Omega(\bm{y}) - (\bm{y} - \mathbf{1}/N)^\top\bm{\theta} \le L_{\Omega}(\bm{\theta}, \mathbf{1}/N) - \Omega(\mathbf{1}/N) - (\bm{y} - \mathbf{1}/N)^\top\bm{\theta},$ due to the assumptions 1--3 which ensure $\Omega$ is non-positive. That is, $L_\Omega(\bm{\theta}, \mathbf{1}/N) \ge \Omega(\mathbf{1}/N) + (\bm{y} - \mathbf{1}/N)^\top\bm{\theta}.$
    Therefore,
    with $\bm{q} = \bm{X}^\top \bm{y}$,
    we get $$E_{\mathrm{concave}}(\bm{q}) \le -\beta^{-1}\Omega(\mathbf{1}/N) - \bm{y}^\top \bm{X} \bm{q} + \bm{q}^\top \bm{\mu}_{\bm{X}} = -\beta^{-1}\Omega(\mathbf{1}/N) - \|\bm{q}\|^2 + \bm{q}^\top \bm{\mu}_{\bm{X}},$$ and
    $E(\bm{q}) = E_{\mathrm{concave}}(\bm{q}) + E_{\mathrm{convex}}(\bm{q}) \le -\beta^{-1}\Omega(\mathbf{1}/N) - \|\bm{q}\|^2 + \bm{q}^\top \bm{\mu}_{\bm{X}} + \frac{1}{2} \|\bm{q} - \bm{\mu}_{\bm{X}}\|^2 + \frac{1}{2}(M^2 - \|\bm{\mu}_{\bm{X}}\|^2) = -\beta^{-1}\Omega(\mathbf{1}/N) -\frac{1}{2}\|\bm{q}\|^2 + \frac{1}{2}M^2 \le -\beta^{-1}\Omega(\mathbf{1}/N) + \frac{1}{2}M^2$.

    To show the second upper bound, use the fact that $E_\mathrm{concave}(\bm{q}) \le 0$, which leads to $E(\bm{q}) \le E_\mathrm{convex}(\bm{q}) = \frac{1}{2}\|\bm{q} - \bm{\mu}_{\bm{X}}\|^2 + \frac{1}{2}(M^2 - \|\bm{\mu}_{\bm{X}}\|^2) = \frac{1}{2}\|\bm{q}\|^2 - \bm{q}^\top\bm{\mu}_{\bm{X}} + \frac{1}{2}M^2$.
    Note that $\|\bm{q}\| = \|\bm{X}^\top \bm{y}\| \le \sum_i y_i \|x_i\| \le M$ and that, from the Cauchy-Schwarz inequality, we have $- \bm{q}^\top\bm{\mu}_{\bm{X}} \le \|\bm{\mu}_{\bm{X}}\| \|\bm{q}\| \le M^2$. Therefore, we obtain $E(\bm{q}) \le \frac{1}{2}\|\bm{q}\|^2 - \bm{q}^\top\bm{\mu}_{\bm{X}} + \frac{1}{2}M^2 \le \frac{1}{2}M^2 +M^2 + \frac{1}{2}M^2 = 2M^2$.


We now turn to the update rule.
The CCCP algorithm works as follows: at the $t$\textsuperscript{th} iteration, it linearizes the concave function $E_{\mathrm{concave}}$ by using a first-order Taylor approximation around $\bm{q}^{(t)}$, $${E}_{\mathrm{concave}}(\bm{q}) \approx \tilde{E}_{\mathrm{concave}}(\bm{q}) := {E}_{\mathrm{concave}}(\bm{q}^{(t)}) + \left(\frac{\partial E_{\mathrm{concave}}(\bm{q}^{(t)})}{\partial \bm{q}}\right)^\top (\bm{q} - \bm{q}^{(t)}).$$
Then, it computes a new iterate by solving the convex optimization problem $\bm{q}^{(t+1)} := \arg\min_{\bm{q}} E_{\mathrm{convex}}(\bm{q}) + \tilde{E}_{\mathrm{concave}}(\bm{q})$, which leads to the equation $\nabla E_\mathrm{convex}(\bm{q}^{(t+1)}) = -\nabla E_\mathrm{concave}(\bm{q}^{(t)})$. Using the fact that $\nabla L_\Omega(\bm{\theta}, \bm{y}) = \hat{\bm{y}}_\Omega(\bm{\theta}) - \bm{y}$ and the chain rule leads to
\begin{align}\label{eq:energy_gradients}
\nabla E_{\mathrm{concave}}(\bm{q}) &= -\beta^{-1} \nabla_{\bm{q}} L_\Omega(\beta\bm{X}\bm{q}; \mathbf{1}/N) = \bm{X}^\top (\mathbf{1}/N - \hat{\bm{y}}_\Omega(\beta\bm{X}\bm{q})) \nonumber\\
&= \bm{\mu}_{\bm{X}} - \bm{X}^\top\hat{\bm{y}}_\Omega(\beta\bm{X}\bm{q})\nonumber\\
\nabla E_{\mathrm{convex}}(\bm{q}) &= \bm{q} - \bm{\mu}_{\bm{X}},
\end{align}
giving the update equation \eqref{eq:updates_hfy}.

\subsection{Proof of Proposition~\ref{prop:separation}}\label{sec:proof_prop_separation}

A stationary point is a solution of the equation $-\nabla E_{\mathrm{concave}}(\bm{q}) = \nabla E_{\mathrm{convex}}(\bm{q})$.
Using the expression for gradients \eqref{eq:energy_gradients},  this is equivalent to $\bm{q} = \bm{X}^\top \hat{\bm{y}}_\Omega(\beta\bm{X}\bm{q})$. If $\bm{x}_i = \bm{X}^\top \bm{e}_i$ is not a convex combination of the other memory patterns,
\(\bm{x}_i\) is a stationary point iff $\hat{\bm{y}}_\Omega(\beta\bm{X}\bm{x}_i)= \bm{e}_i$.
We now use the margin property of sparse transformations \eqref{eq:margin}, according to which the latter is equivalent to $\beta\bm{x}_i^\top \bm{x}_i - \max_{j\ne i} \beta\bm{x}_i^\top \bm{x}_j \ge m$. 
Noting that the left hand side equals $\beta\Delta_i$ leads to the desired result. 

If the initial query satisfies ${\bm{q}^{(0)}}^\top (\bm{x}_i - \bm{x_j}) \ge \frac{m}{\beta}$ for all $j \ne i$, we have again from the margin property that $\hat{\bm{y}}_\Omega(\beta\bm{X}\bm{q}^{(0)})= \bm{e}_i$, which combined to the previous claim ensures convergence in one step to $\bm{x}_i$. 

Finally, note that, 
if $\bm{q}^{(0)}$ is $\epsilon$-close to $\bm{x}_i$, we have 
$\bm{q}^{(0)} = \bm{x}_i + \epsilon \bm{r}$ for some vector $\bm{r}$ with $\|\bm{r}\|=1$. 
Therefore, we have 
\begin{align}
(\bm{q}^{(0)})^\top (\bm{x}_i - \bm{x}_j) &= (\bm{x}_i + \epsilon \bm{r})^\top (\bm{x}_i - \bm{x}_j) \nonumber\\
&\ge \Delta_i + \epsilon \bm{r}^\top (\bm{x}_i - \bm{x}_j) \nonumber\\
&\ge \Delta_i - \epsilon \underbrace{\|\bm{r}\|}_{=1}  \|\bm{x}_i - \bm{x}_j\|,
\end{align}
where we invoked the Cauchy-Schwarz inequality in the last step. 
Since the patterns are normalized (with norm $M$),%
\footnote{In fact, the result still holds if patterns are not normalized but have their norm upper bounded by $M$, i.e., if they lie within a ball of radius $M$ and not necessarily on the sphere.} %
we have from the triangle inequality that $\|\bm{x}_i - \bm{x}_j\| \le \|\bm{x}_i\| + \|\bm{x}_j\| = 2M$; using the assumption that $\Delta_i \ge \frac{m}{\beta} + 2M\epsilon$, we obtain ${\bm{q}^{(0)}}^\top (\bm{x}_i - \bm{x}_j) \ge \frac{m}{\beta}$, which from the previous points ensures convergence to $\bm{x}_i$ in one iteration.

\subsection{Proof of Proposition~\ref{prop:storage}}~\label{sec:proof_prop_storage}

For the first statement, we follow a similar argument as the one made by \citet{ramsauer2020hopfield} in the proof of their Theorem A.3---however their proof has a mistake, which we correct here.%
\footnote{Concretely, \citet{ramsauer2020hopfield} claim that given a separation angle $\alpha_{\min}$, we can place $N = \left({2\pi}/{\alpha_{\min}}\right)^{D-1}$ patterns equidistant on the sphere, but this is not correct.} %
Given a separation angle $\alpha_{\min}$, we lower bound the number of patterns $N$ we can can place in the sphere separated by at least this angle. 
Estimating this quantity is an important open problem in combinatorics, related to determining the size of spherical codes (of which kissing numbers are a particular case; \citealt{conway2013sphere}). We invoke a lower bound due to \citet{chabauty1953resultats}, \citet{shannon1959probability}, and \citet{wyner1965capabilities} (see also \citet{jenssen2018kissing} for a tighter bound), who show that $N \ge (1 + o(1))\sqrt{2\pi D} \frac{\cos \alpha_{\min}}{(\sin \alpha_{\min})^{D-1}}$. 
For $\alpha_{\min} = \frac{\pi}{3}$, which corresponds to the kissing number problem, we obtain the bound $$N \ge (1 + o(1))\sqrt{\frac{3\pi D}{8}}\left(\frac{2}{\sqrt{3}}\right)^D = \mathcal{O}\left(\left(\frac{2}{\sqrt{3}}\right)^D\right).$$
In this scenario, we have $\Delta_i = M^2(1 - \cos \alpha_{\min})$ by the definition of $\Delta_i$. 
From Proposition~\ref{prop:separation}, we have exact retrieval under $\epsilon$-perturbations if $\Delta_i \ge m\beta^{-1} + 2M\epsilon$. 
Combining the two expressions, we obtain $\epsilon \le \frac{M}{2}(1 - \cos \alpha_{\min}) - \frac{m}{2\beta M}$. 
Setting $\alpha_{\min} = \frac{\pi}{3}$, we obtain
$\epsilon \le \frac{M}{2}\left(1 - \frac{1}{2}\right) - \frac{m}{2\beta M} = \frac{M}{4} - \frac{m}{2\beta M}$. 
For the right hand side to be positive, we must have $M^2 > 2m/\beta$.


Assume now patterns are placed uniformly at random in the sphere. From \citet{brauchart2018random} we have, for any $\delta>0$:  
\begin{align}
    P(N^{\frac{2}{D-1}}\alpha_{\min} \ge \delta) \ge 1 - \frac{\kappa_{D-1}}{2}\delta^{D-1}, \quad \text{with} \quad
    \kappa_{D} := \frac{1}{D\sqrt{\pi}}\frac{\Gamma((D+1)/2)}{\Gamma(D/2)}.
\end{align}
Given our failure probability $p$, we need to have 
\begin{align}
    P(M^2 (1 - \cos \alpha_{\min}) \ge m\beta^{-1} + 2M\epsilon) \ge 1-p,
\end{align}
which is equivalent to
\begin{align}
P \left\{ N^{\frac{2}{D-1}} \alpha_{\min} \ge \underbrace{N^{\frac{2}{D-1}} \arccos \left( 1 - \frac{m}{\beta M^2} - \frac{2\epsilon}{M} \right)}_{:= \delta} \right\} \ge 1-p.
\end{align}
Therefore, we set 
\begin{align}
    p = \frac{\kappa_{D-1}}{2} \delta^{D-1} = \frac{\kappa_{D-1}}{2} N^2 \left[ \arccos \left( 1 - \frac{m}{\beta M^2} - \frac{2\epsilon}{M} \right)\right]^{D-1}.
\end{align}
Choosing $N =  \sqrt{\frac{2p}{\kappa_{D-1}}} \zeta^{\frac{D-1}{2}}$ patterns for some $\zeta > 1$, we obtain 
\begin{align}
    1 = \left[\zeta \arccos \left( 1 - \frac{m}{\beta M^2} - \frac{2\epsilon}{M} \right)\right]^{D-1}. 
\end{align}
Therefore the failure rate $p$ is attainable provided the perturbation error is 
\begin{equation}
    \epsilon \le  \frac{M}{2} \left(1 - \cos \frac{1}{\zeta}\right) - \frac{m}{2\beta M}.
\end{equation}
For the right hand side to be positive, we must have
$\cos \frac{1}{\zeta} < 1 - \frac{m}{\beta M^2}$, i.e., $\zeta < \frac{1}{\arccos \left(1 -  \frac{m}{\beta M^2}\right)}$. 

\subsection{Proof of Proposition~\ref{prop:sparsemap_margin}}\label{sec:proof_prop_sparsemap_margin}

The first statement in the proposition is stated and proved by \citet{blondel2020learning} as a corollary of their Proposition 8. 
We prove here a more general version, which includes the second statement as a novel result. 

Using \citet[Proposition 8]{blondel2020learning}, we have that the structured margin of $L_\Omega$ is given by the following expression,
\begin{align*}
    m = \sup_{\substack{\bm{y} \in \mathcal{Y}\\\bm{\mu} \in \mathrm{conv}(\mathcal{Y})}} \frac{\Omega(\bm{y}) - \Omega(\bm{\mu})}{r^2 - \bm{\mu}^\top \bm{y}},
\end{align*}
if the supremum exists. 
For SparseMAP, using $\Omega(\bm{\mu}) = \frac{1}{2}\|\bm{\mu}_V\|^2 = \frac{1}{2}\|\bm{\mu}\|^2 - \frac{1}{2}\|\bm{\mu}_F\|^2$ for any $\bm{\mu} \in \mathrm{conv}(\mathcal{Y})$, and using the fact that $\|\bm{y}\| = r$ for any $\bm{y} \in \mathcal{Y}$, we obtain:
\begin{align*}
    m &= \sup_{\substack{\bm{y} \in \mathcal{Y}\\\bm{\mu} \in \mathrm{conv}(\mathcal{Y})}} \frac{\frac{1}{2}r^2 - \frac{1}{2}\|\bm{\mu}\|^2 +  \overbrace{\frac{1}{2}\|\bm{\mu}_F\|^2 - \frac{1}{2}r_F^2}^{\le 0}}{\bm{y}^\top (\bm{y} - \bm{\mu})} \nonumber\\
    &
    \le^{(\dagger)} \sup_{\substack{\bm{y} \in \mathcal{Y}\\\bm{\mu} \in \mathrm{conv}(\mathcal{Y})}} \frac{\frac{1}{2}r^2 - \frac{1}{2}\|\bm{\mu}\|^2}{\bm{y}^\top (\bm{y} - \bm{\mu})}\nonumber\\
    &= 1 - \inf_{\substack{\bm{y} \in \mathcal{Y}\\\bm{\mu} \in \mathrm{conv}(\mathcal{Y})}} \frac{\frac{1}{2}\|\bm{y} - \bm{\mu}\|^2}{\bm{y}^\top (\bm{y} - \bm{\mu})}\nonumber\\
    &\le^{(\ddagger)} 1,
\end{align*}
where the inequality $^{(\dagger)}$ follows from the convexity of $\frac{1}{2}\|\cdot\|^2$, which implies that $\frac{1}{2}\|\bm{\mu}_F\|^2 \le \frac{1}{2}\|\bm{y}_F\|^2 = \frac{1}{2}r_F^2$; and the inequality $^{(\ddagger)}$ follows from the fact that both the numerator and denominator in the second term are non-negative, the latter due to the Cauchy-Schwartz inequality and the fact that $\|\bm{\mu}\| \le r$. 
This proves the second part of Proposition~\ref{prop:sparsemap_margin}. 

To prove the first part, note first that,  
if there are no higher order interactions, then $r_F = 0$ and $\bm{\mu}_F$ is an ``empty vector'', which implies that $^{(\dagger)}$ is an equality. 
We prove now that, in this case, $^{(\ddagger)}$ is also an equality, which implies that $m=1$. 
We do that by showing that, 
for any $\bm{y} \in \mathcal{Y}$, we have $\inf_{\bm{\mu} \in \mathrm{conv}(\mathcal{Y})} \frac{\frac{1}{2}\|\bm{y}-\bm{\mu}\|^2}{\bm{y}^\top (\bm{y} - \bm{\mu})} = 0$. 
Indeed, choosing $\bm{\mu}= t\bm{y}' + (1-t)\bm{y}$ for an arbitrary $\bm{y}' \in \mathcal{Y} \setminus \{\bm{y}\}$, and letting $t\rightarrow 0^+$, we obtain $\frac{\frac{1}{2}\|\bm{y}-\bm{\mu}\|^2}{\bm{y}^\top (\bm{y} - \bm{\mu})} = \frac{\frac{t}{2}\|\bm{y}-\bm{y}'\|^2}{\bm{y}^\top (\bm{y} - \bm{y}')} \rightarrow 0$.

\subsection{Proof of Proposition~\ref{prop:stationary_single_iteration_sparsemap}}\label{sec:proof_prop_stationary_single_iteration_sparsemap}

A point $\bm{q}$ is stationary iff it satisfies $\bm{q} = \bm{X}^\top \hat{\bm{y}}_\Omega(\beta \bm{X}\bm{q})$.  
Therefore,   
$\bm{X}^\top \bm{y}_i$ is guaranteed to be a stationary point if%
\footnote{But not necessarily ``only if'' -- in general, we could have $\bm{X}^\top\bm{y}_i$ in the convex hull of the other pattern associations.} %
$\hat{\bm{y}}_\Omega(\beta \bm{X}\bm{X}^\top \bm{y}_i) = \bm{y}_i$, which is equivalent to zero loss, i.e., to the existence of a margin 
$\underbrace{\beta \bm{y}_i^\top \bm{X}\bm{X}^\top (\bm{y}_i - \bm{y}_j)}_{\ge \beta \Delta_i} \ge \frac{1}{2}\|\bm{y}_i - \bm{y}_j\|^2$ for all $j$. 
Since we are assuming $\Delta_i \ge \frac{D_i^2}{2\beta} \ge \frac{\|\bm{y}_i - \bm{y}_j\|^2}{2\beta}$ for all $j$, we have $\beta \Delta_i \ge \frac{\|\bm{y}_i - \bm{y}_j\|^2}{2}$ for all $j$, which implies the margin condition above. 

If the initial query satisfies $\bm{q}^\top 
\bm{X}^\top(\bm{y}_i - \bm{y}_j) \ge \frac{D_i^2}{2\beta}$ for all $j \ne i$, we have again from the margin property that $\hat{\bm{y}}_\Omega(\beta\bm{X}\bm{q})= \bm{y}_i$, which ensures convergence in one step to $\bm{X}^\top\bm{y}_i$. 

If $\bm{q}$ is $\epsilon$-close to $\bm{X}^\top\bm{y}_i$, then we have 
$\bm{q} = \bm{X}^\top\bm{y}_i + \epsilon \bm{r}$ for some vector $\bm{r}$ with $\|\bm{r}\|=1$. 
Therefore, we have 
\begin{align}
\bm{q}^\top \bm{X}^\top (\bm{y}_i - \bm{y}_j) &= (\bm{X}^\top\bm{y}_i + \epsilon \bm{r})^\top \bm{X}^\top (\bm{y}_i - \bm{y}_j) \nonumber\\
&\ge \Delta_i + \epsilon \bm{r}^\top \bm{X}^\top (\bm{y}_i - \bm{y}_j).  \end{align}
We now bound $-\bm{r}^\top \bm{X}^\top (\bm{y}_i - \bm{y}_j)$ in two possible ways. 
Using the Cauchy-Schwarz inequality, 
we have $-\bm{r}^\top \bm{X}^\top (\bm{y}_i - \bm{y}_j) \le \|\bm{X} \bm{r}\| \|\bm{y}_i - \bm{y}_j\| \le \sigma_{\max}(\bm{X}) D_i$, where $\sigma_{\max}(\bm{X})$ denotes the largest singular value of $\bm{X}$, i.e., the spectral norm of $\bm{X}$. 
On the other hand, denoting 
$R_i := \max_j \|\bm{y}_i - \bm{y}_j\|_1$, we can also use H\"older's inequality to obtain $-\bm{r}^\top \bm{X}^\top (\bm{y}_i - \bm{y}_j) \le \|\bm{X} \bm{r}\|_\infty \|\bm{y}_i - \bm{y}_j\|_1 \le M R_i$, where we used the fact that 
$\|\bm{X} \bm{r}\|_\infty = \max_k |\bm{x}_k^\top \bm{r}| \le \|\bm{x}_k\|\|\bm{r}\| = M$. Combining the two inequalities, we obtain 
$\bm{q}^\top \bm{X}^\top (\bm{y}_i - \bm{y}_j) \ge \Delta_i - \epsilon \min \{\sigma_{\max}(\bm{X}) D_i, MR_i\}$. 
Using the assumption that $\Delta_i \ge \frac{D_i^2}{2\beta} + \epsilon\min \{\sigma_{\max}(\bm{X}) D_i, MR_i\}$, we obtain $\bm{q}^\top \bm{X}^\top (\bm{y}_i - \bm{y}_j) \ge \frac{D_i^2}{2\beta}$, which from the previous points ensures convergence to $\bm{X}^\top\bm{y}_i$ in one iteration. 
The result follows by noting that, since $\mathcal{Y} \subseteq \{0,1\}^D$, we have $R_i = D_i^2$.  

\section{Additional Details and Experiments}
\subsection{Sparse and structured transformations}
\label{sec:SST_App}
\begin{figure}[t]
    \centering
    \includegraphics[width=\columnwidth]{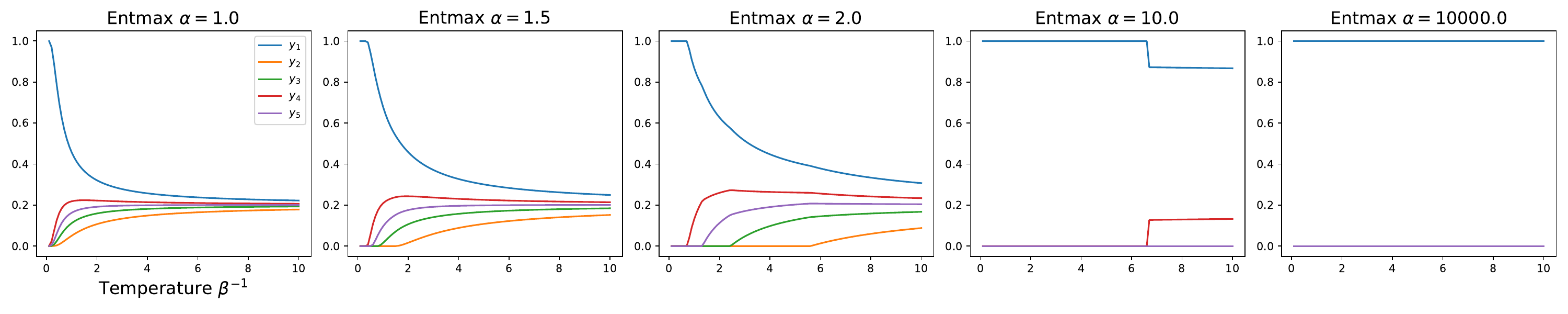}\\
    \vspace{0.3cm}
    \includegraphics[width=\columnwidth]{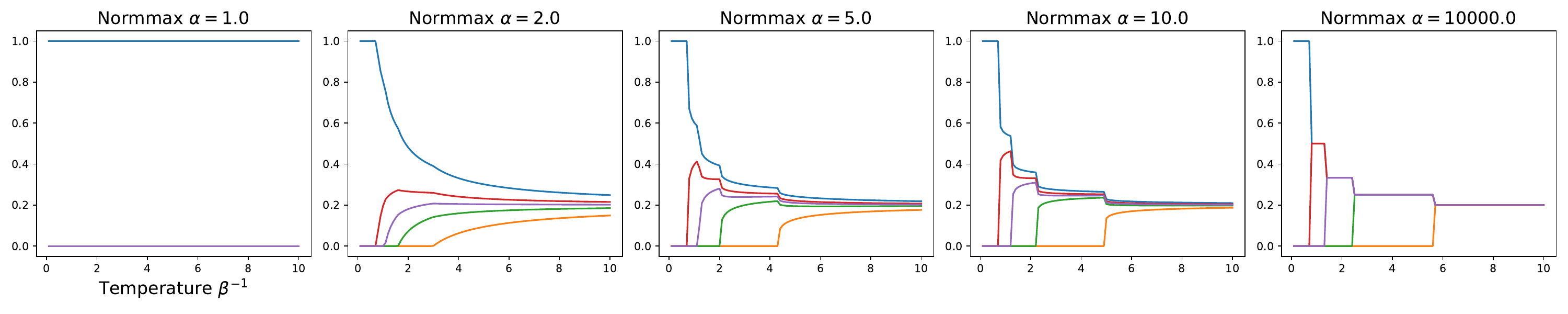}\\
    \vspace{0.3cm}
    \includegraphics[width=\columnwidth]{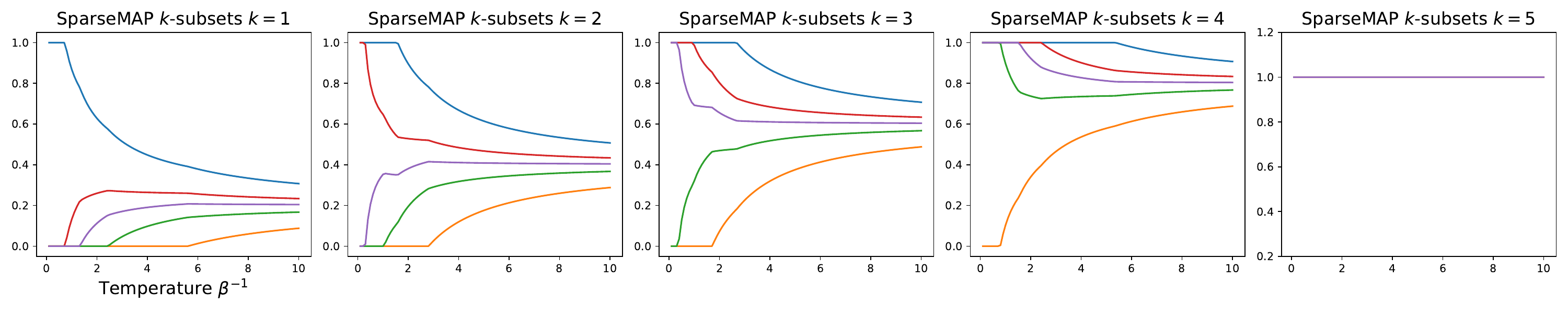}
    \caption{Sparse and structured transformations used in this paper and their regularization path. In each plot, we show $\hat{\bm{y}}_\Omega(\beta \bm{\theta}) = \hat{\bm{y}}_{\beta^{-1}\Omega}(\bm{\theta})$ as a function of the temperature $\beta^{-1}$ where $\bm{\theta} = [1.0716, -1.1221, -0.3288, 0.3368, 0.0425]^\top$.}
    \label{fig:sparse_transformations1}
\end{figure} 
We show additional sparse and structured transformations and their regularization paths in Figure \ref{fig:sparse_transformations1}. The difference between the entmax and normmax regularizers is subtle but important: when $\alpha \rightarrow \infty$, the entmax regularizer vanishes and entmax becomes argmax, returning a one-hot vector. However, when $\alpha \rightarrow \infty$, the normmax regularizer becomes $\Omega_{\infty}^N(\bm{y}) = -1+\max_i (y_i)$ and the transformation returns a uniform distribution with a sparse support, as shown in Figure \ref{fig:sparse_transformations1}, rightmost middle plot. On the other hand, when $\alpha \rightarrow 1$, entmax becomes softmax, but the normmax regularizer vanishes and normmax becomes argmax. SparseMAP with $k$-subsets, as $\beta$ increases,  tends to return a $k$-hot vector, where for $k=1$ it corresponds to entmax with $\alpha=2$ (sparsemax).

\subsection{Hopfield dynamics and basins of attraction}
\label{sec:HDBA}
\begin{figure*}[t]
  \centering
  \includegraphics[width=0.49\textwidth]{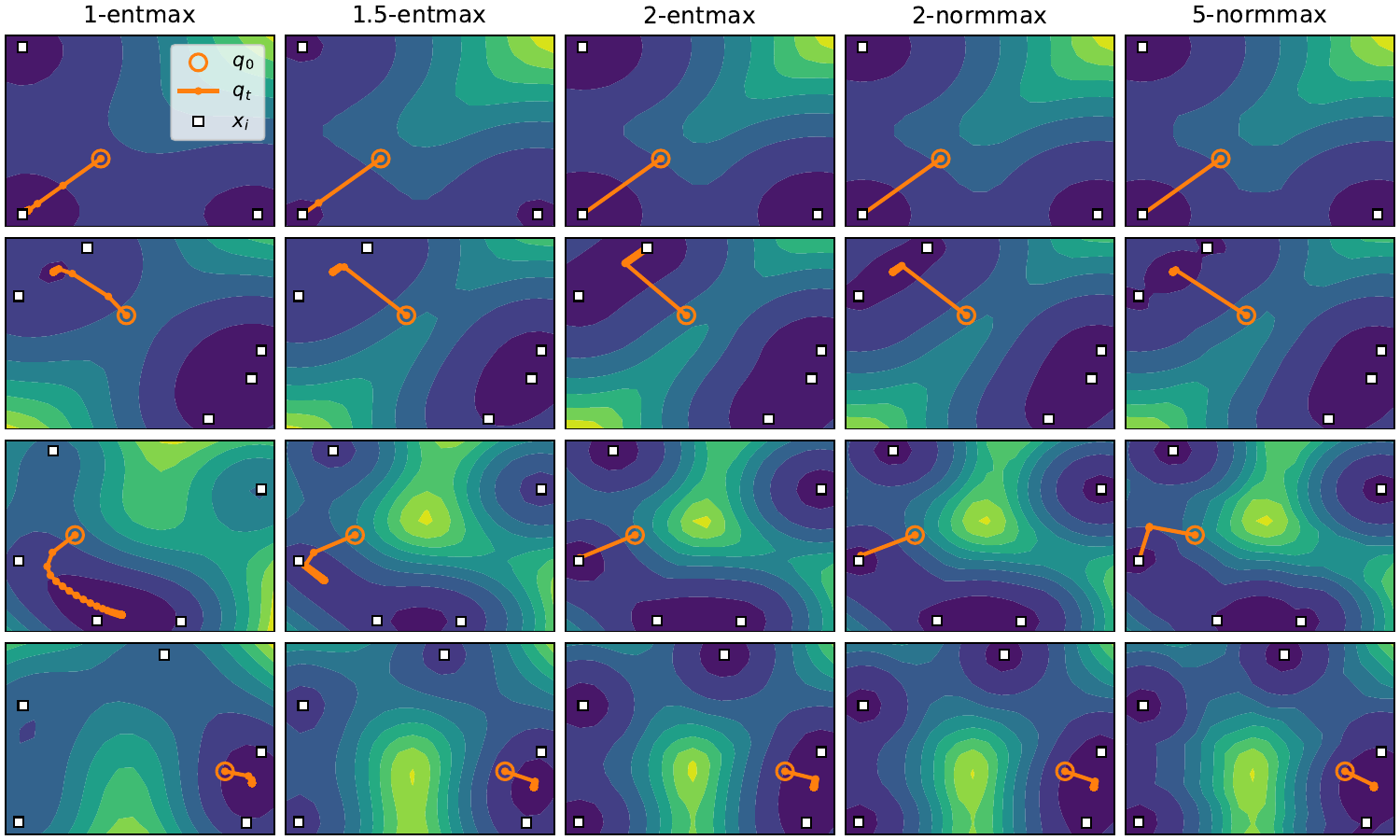}\hspace{2pt}%
  \includegraphics[width=0.49\textwidth]{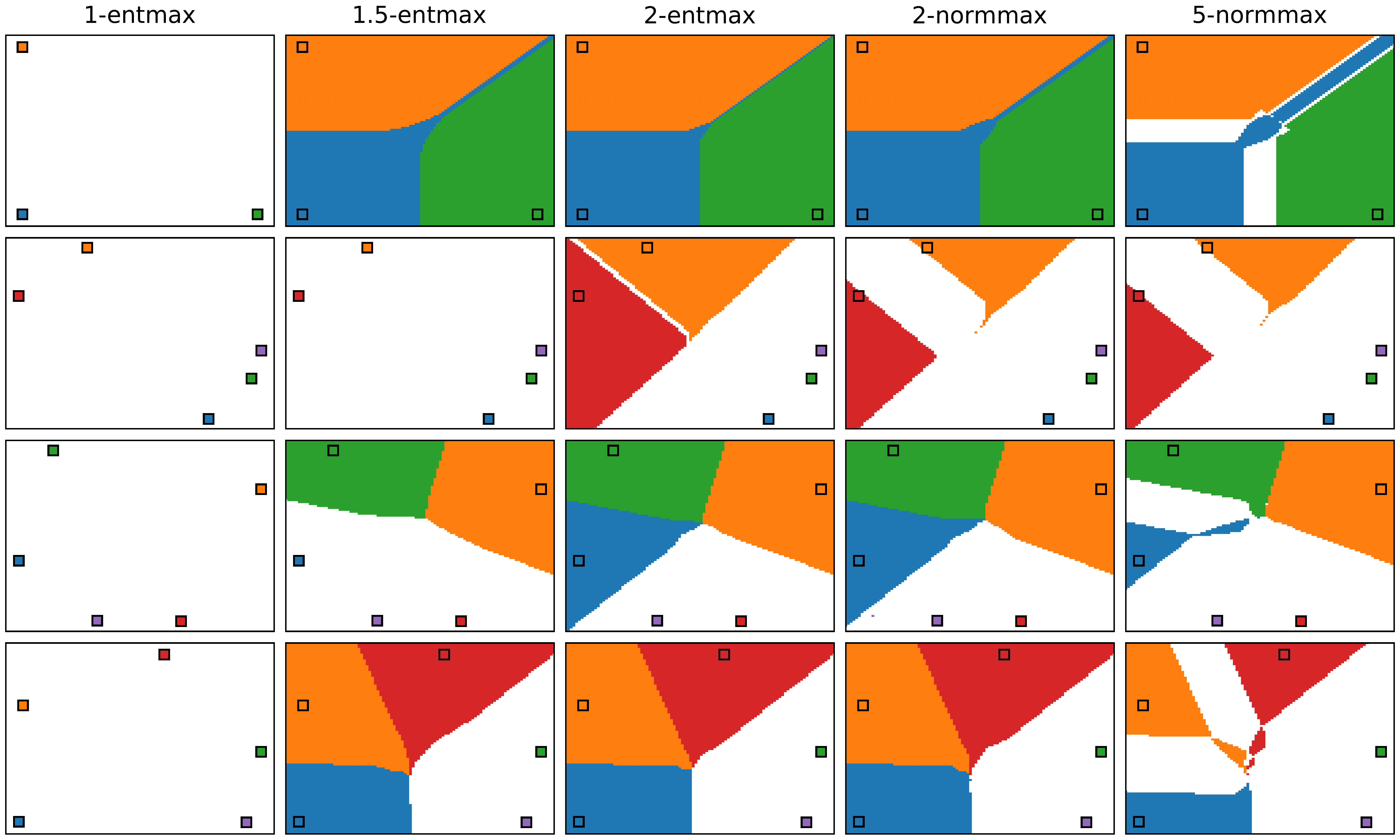} 
\includegraphics[width=0.49\textwidth]{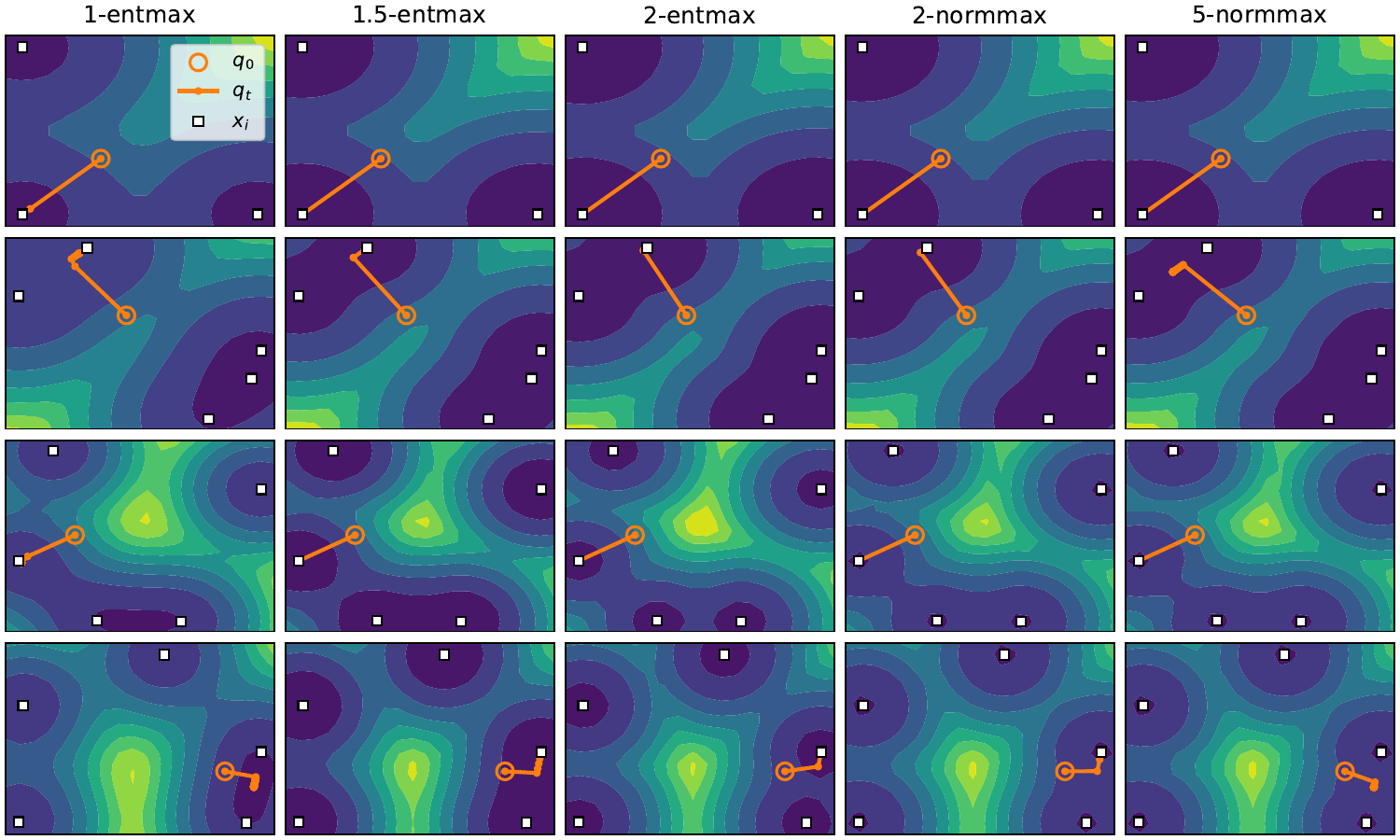}
  \includegraphics[width=0.49\textwidth]{basins.png}
\caption{Top: Contours of the energy function and optimization trajectory of the CCCP iteration (left) alongside the attraction basins associated with each pattern (right) for $\beta=4$. (Note that white sections do not converge to a single pattern but to a metastable state.) Bottom: Similar representation for $\beta=10$.}
  \label{fig:overall1}
\end{figure*}

In Figure \ref{fig:overall1}, additional plots with varying $\beta$ values are provided. Particularly, as $\beta$ increases, the optimization trajectories exhibit a tendency to converge towards a single pattern, contrasting with the prevalence of metastable states observed for smaller $\beta$ values. Moreover, the basins of attraction become increasingly colorful with higher $\beta$ values, suggesting a convergence behaviour similar to what was previously described.
\section{Experimental Details}
\subsection{MNIST $K$-MIL}
\label{sec:MNIST_Experimental_Details}
For $K$-MIL, we created 4 datasets by grouping the MNIST examples into bags, for $K \in \{1, 2, 3, 5\}$. A bag is positive if it contains at least $K$ targets, where the target is the number ``9'' (we chose ``9'' as it can be easily misunderstood with ``7'' or ``4''). The embedding architecture is the same as \citet{ilse2018attention}, but instead of attention-based pooling, we use our $\alpha$-entmax pooling, with $\alpha=1$ mirroring the pooling method in \cite{ramsauer2020hopfield}, and $\alpha=2$ corresponding to the pooling in \cite{hu2023sparse}. Additionally, we incorporate $\alpha$-normmax pooling and SparseMAP pooling with $k$-subsets. Further details of the $K$-MIL datasets are shown in Table~\ref{tab:MNIST_dataset}.
\begin{table}[t]
    \centering
    \caption{Dataset sample details for the MNIST $K$-MIL experiment. The size $L_i$ of the $i\textsuperscript{th}$ bag is determined through $L_i = \max\{K, L_i'\}$ where $L_i' \sim \mathcal{N}(\mu, \sigma^2)$.  
    The number of positive instances in a bag is uniformly sampled between $K$ and $L_i$ for positive bags and between $0$ and $K-1$ for negative bags.}
    \label{tab:MNIST_dataset}
    \begin{tabular}{cccccc}
        \toprule
        {\textbf{Dataset}} & {$\boldsymbol{\mu}$} & {$\boldsymbol{\sigma}$} & {\textbf{Features}} & {\textbf{Pos. training bags}} & {\textbf{Neg. training bags}} \\
        \midrule
        MNIST, $K=1$ & 10 & 1 & 28 $\times$ 28 & 1000 & 1000 \\
        MNIST, $K=2$ & 11 & 2 & 28 $\times$ 28 & 1000 & 1000 \\
        MNIST, $K=3$ & 12 & 3 & 28 $\times$ 28 & 1000 & 1000\\
        MNIST, $K=5$ & 14 & 5 & 28 $\times$ 28 & 1000 & 1000 \\
        \bottomrule
    \end{tabular}
\end{table}

We train the models for 5 different random seeds, where the first one is used for tuning the hyperparameters. The reported test accuracies represent the average across these seeds. We use 500 bags for testing and 500 bags for validation. The hyperparameters are tuned via grid search, where the grid space is shown in Table \ref{tab:MNIST_hyperparam}. We consider a dropout hyperparameter, commonly referred to as bag dropout, to the Hopfield matrix due to the risk of overfitting (as done by \citet{ramsauer2020hopfield}). 
All models were trained for 50 epochs. We incorporated an early-stopping mechanism, with patience 5, that selects the optimal checkpoint based on performance on the validation set. 

\begin{table}[t]
    \centering
    \caption{Hyperparameter space for the MNIST MIL experiment. Hidden size is the dimension of keys and queries and $\gamma$ is a parameter of the exponential learning rate scheduler \citep{Li2020An}.}
    \label{tab:MNIST_hyperparam}
    \begin{tabular}{ l  l }
        \toprule
        {\textbf{Parameter}} & {\textbf{Range}} \\
        \midrule
        learning rate & \{$10^{-5}$, $10^{-6}$\} \\
        $\gamma$ & \{0.98 , 0.96\}  \\
        hidden size & \{16, 64\} \\
        number of heads & \{8, 16\} \\
        $\beta$ & \{0.25, 0.5, 1.0, 2.0, 4.0, 8.0\} \\
        bag dropout & \{0.0, 0.75\} \\
        \bottomrule
    \end{tabular}
\end{table}

\subsection{MIL benchmarks}
\label{sec:MIL_bench_details}
The MIL benchmark datasets (Fox, Tiger and Elephant) comprise preprocessed and segmented color images sourced from the Corel dataset \cite{ilse2018attention}. Each image is composed of distinct segments or blobs, each defined by descriptors such as color, texture, and shape. The datasets include 100 positive and 100 negative example images, with the negative ones randomly selected from a pool of photos featuring various other animals.

The $\mathrm{HopfieldPooling}$ layers ($\alpha$-entmax; $\alpha$-normmax; SparseMAP, $k$-subsets) take as input a collection of embedded instances, along with a trainable yet constant query. This query pattern is used for the purpose of averaging class-indicative instances, thereby facilitating the compression of bags of variable sizes into a consistent representation. This compression is important for effectively discriminating between different bags. To tune the model, a manual hyperparameter search was conducted on a validation set.

In our approach to tasks involving Elephant, Fox and Tiger, we followed a similar architecture as \citep{ramsauer2020hopfield}:
\begin{enumerate}
    \item The first two layers are fully connected linear embedding layers with ReLU activation.
    \item The output of the second layer serves as the input for the $\mathrm{HopfieldPooling}$  layer, where the pooling operation is executed.
    \item Subsequently, we employ a single layer as the final linear output layer for classification with a sigmoid as the classifier.
\end{enumerate}

During the hyperparameter search, various configurations were tested, including different hidden layer widths and learning rates. Particular attention was given to the hyperparameters of the $\mathrm{HopfieldPooling}$ layers, such as the number of heads, head dimension, and the inverse temperature $\beta$. To avoid overfitting, bag dropout (dropout at the attention weights) was implemented as the chosen regularization technique. All hyperparameters tested are shown in Table \ref{tab:fox_hyperparam}.
\begin{table}[h]
    \centering
    \caption{Hyperparameter space for the MIL benchmark experiments. Hidden size is the space in which keys and queries are associated and $\gamma$ is a parameter of the exponential learning rate scheduler.}
    \label{tab:fox_hyperparam}
    \begin{tabular}{ l  l }
        \toprule
        {\textbf{Parameter}} & {\textbf{Range}} \\
        \midrule
        learning rate & \{$10^{-3}$, $10^{-5}$\} \\
        $\gamma$ & \{0.98 , 0.96\}  \\
        embedding dimensions & \{32 , 128\}  \\
        embedding layers & \{2\}  \\
        hidden size & \{32, 64\} \\
        number of heads & \{12\} \\
        $\beta$ & \{0.1, 1, 10\} \\
        bag dropout & \{0.0, 0.75\} \\
        \bottomrule
    \end{tabular}
\end{table}
We trained for 50 epochs with early stopping with patience 5, using the Adam optimizer \cite{loshchilov2017decoupled} with exponential learning rate decay. Model validation was conducted through a 10-fold nested cross-validation, repeated five times with different data splits where the first seed is used for hyperparameter tuning. The reported test ROC AUC scores represent the average across these repetitions.

\subsection{Text Rationalization}\label{sec:text_rationalization_details}
We use the same hyperparameters reported by \citet{guerreiro2021spectra}. 
We used a head dimension of 200, to match the dimensions of the encoder vectors  (the size of the projection matrices associated to the static query and keys) and a head dropout of 0.5 (applied to the output of the Hopfield layer). We used a single attention head to better match the SPECTRA model.  Aditionally we use a transition score of 0.001 and a train temperature of 0.1.

\begin{figure*}[b]
    \centering
    \includegraphics[width=1\textwidth]{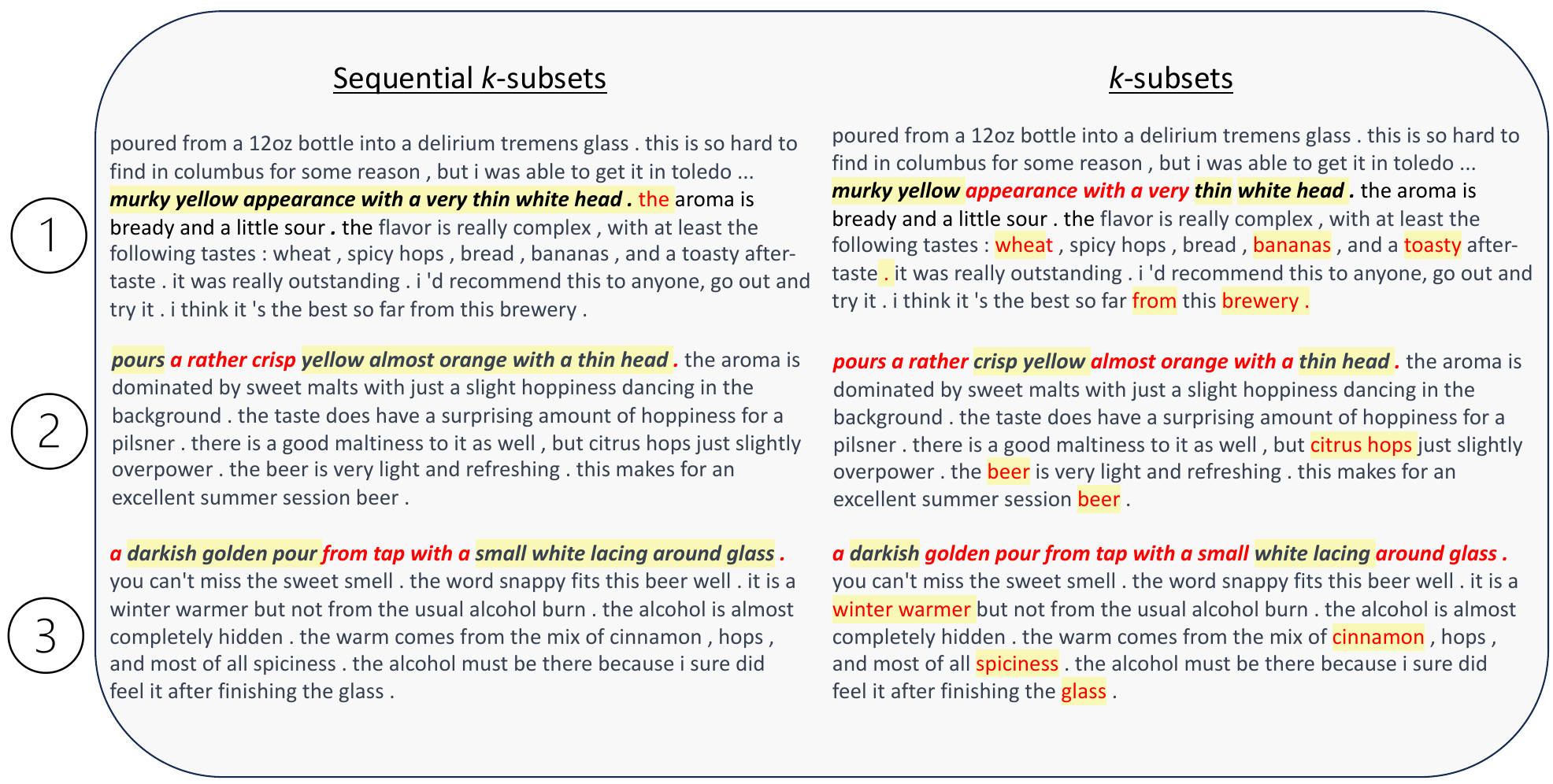}
    \caption{Examples of human rationale overlap for the aspect ``appearance''. The \hl{yellow highlight} indicates the model's rationale, while \textbf{\textit{italicized and bold font}} represents the human rationale. \textcolor{red}{Red font} identifies mismatches with human annotations. SparseMAP  with sequential $k$-subsets prefers more contiguous  rationales, which better match humans.}
    \label{fig:mismatch}
\end{figure*}

We present in Table \ref{tab:spectra_extended} an extended version of Table \ref{tab:spectra} with additional baselines, corresponding to prior work in text rationalization. 
We show additional examples of rationales extracted by our models in Figure~\ref{fig:mismatch}.  
\begin{table*}[t]
\caption{Text rationalization results. We report mean and min/max $F_1$ scores across ﬁve random seeds on test sets for all datasets but Beer, where we report
MSE. All entries except SparseMAP are taken from \cite{guerreiro2021spectra}. We also report human rationale overlap (HRO) as $F_1$ score. We bold the best-performing rationalized model(s).}
\centering
\small
\begin{tabular}{l l cccccc}
\toprule
Method & Rationale  & SST$\uparrow$ & AgNews$\uparrow$ & IMDB$\uparrow$ & Beer$\downarrow$  & Beer(HRO)$\uparrow$\\
\midrule
\multirow{2}{*}{SFE} & top-$k$ & .76 {\small(.71/.80)} & .92 {\small(.92/.92)} & .84 {\small(.72/.88)} & .018 {\small(.016/.020)} & .19 {\small(.13/.30)} \\
& contiguous & .71 {\small(.68/.75)} & .86 {\small(.85/.86)}& .65 {\small(.57/.73)} &.020 {\small(.019/.024)} & 35 {\small(.18/.42)} \\
\midrule
\multirow{2}{*}{SFE w/Baseline} & top-$k$ & .78 {\small(.76/.80)} & .92 {\small(.92/.93)} & .82 {\small(.72/.88)} & .019 {\small(.017/.020)} & .17 {\small(.14/.19)}  \\
& contiguous & .70 {\small(.64/.75)} & .86 {\small(.84/.86)}& .76 {\small(.73/.80)} &.021 {\small(.019/.025)} & {.41\small(.37/.42)} \\
\midrule
\multirow{2}{*}{Gumbel} & top-$k$ & .70 {\small(.67/.72)} & .78 {\small(.73/.84)} & .74 {\small(.71/.78)} & .026 {\small(.018/.041)} & .27 {\small(.14/.39)} \\
& contiguous & .67 {\small(.67/.68)}& .77 {\small(.74/.81)} & .72 {\small(.72/.73)} & .043 {\small(.040/.048)} & .42 {\small(.41/.42)}\\
\midrule
HardKuma & - & .80 {\small(.80/.81)} & .90 {\small(.87/.88)} & .87 {\small(.90/.91)} & .019 {\small(.016/.020)} & .37 {\small(.00/.90)}  \\
\midrule
\multirow{2}{*}{Sparse Attention} & sparsemax & \textbf{.82} {\small(.81/.83)} & \textbf{.93} {\small(.93/.93)} & .89 {\small(.89/.90)} & .019 {\small(.016/.021)} & 48 {\small(.41/.55)}  \\
& fusedmax & .81 {\small(.81/.82)}& .92 {\small(.91/.92)}& .88 {\small(.87/.89)} & .018 {\small(.017/.019)} & 39 {\small(.29/.53)} \\
\midrule
SPECTRA & sequential $k$-subsets & .80 {\small(.79/.81)} & .92 {\small(.92/.93)} & \textbf{.90} {\small(.89/.90)} & \textbf{.017} {\small(.016/.019)} & .61 {\small(.56/.68)}  \\
\midrule
\multirow{2}{*}{SparseMAP} &  $k$-subsets & .81 {\small(.81/.82)}  & \textbf{.93} {\small(.92/.93)} & \textbf{.90} {\small(.90/.90)} & \textbf{.017} {\small(.017/.018)} & .42 {\small(.29/.62)}   \\
& sequential $k$-subsets & .81 {\small(.80/.83)}& \textbf{.93} {\small(.93/.93)} & \textbf{.90} {\small(.90/.90)} & .020 {\small(.018/.021)} & \textbf{.63} {\small(.49/.70)} \\
\bottomrule
\end{tabular}
\label{tab:spectra_extended}
\end{table*}

\end{document}